\documentclass[acmsmall]{acmart}

\AtBeginDocument{%
  }

\setcopyright{acmcopyright}
\copyrightyear{2024}
\acmYear{2024}
\acmDOI{XXXXXXX.XXXXXXX}

\usepackage{algorithm}
\usepackage[noend]{algpseudocode}
\DeclareMathOperator*{\argmax}{argmax}
\acmJournal{JACM}
\acmVolume{0}
\acmNumber{0}
\acmArticle{0}
\acmMonth{4}

\newcommand\restr[2]{{%
  \left.\kern-\nulldelimiterspace %
  #1 %
  \right|_{#2} %
  }}

\newcommand{\rem}[1]{}
\newcommand\modif[1]{
\color{blue}
#1
\color{black}
}

\begin{document}

\title{4D Facial Expression Diffusion Model}

\author{Kaifeng Zou}
\affiliation{%
  \institution{ICube Laboratory, University of Strasbourg}
  \streetaddress{300 bd Sébastien Brant}
  \city{Illkirch}
  \country{France}
  \postcode{67402}
}
\email{kaifeng.zou@unistra.fr}
\orcid{0000-0003-4460-3690}

\author{Sylvain Faisan}
\email{faisan@unistra.fr}
\affiliation{%
  \institution{ICube Laboratory, University of Strasbourg}
  \streetaddress{300 bd Sébastien Brant}
  \city{Illkirch}
  \country{France}
  \postcode{67402}
}

\author{Boyang Yu}
\affiliation{%
  \institution{ICube Laboratory, University of Strasbourg}
  \streetaddress{1 Pl. de l'Hôpital}
  \city{Strasbourg}
  \country{France}}
\email{b.yu@unistra.fr}

\author{S\'ebastien Valette} 
\affiliation{%
  \institution{CREATIS, CNRS, INSA-Lyon, Lyon, France}
  \streetaddress{21 Av. Jean Capelle O}
  \city{Villeurbanne}
  \country{France}
}
\email{sebastien.valette@creatis.insa-lyon.fr}

\author{Hyewon Seo}
\affiliation{%
 \institution{ICube Laboratory, University of Strasbourg}
 \streetaddress{1 Pl. de l'Hôpital}
 \city{Strasbourg}
 \country{France}}
 \email{seo@unistra.fr}

\renewcommand{\shortauthors}{Zou et al.}

\begin{abstract}
 Facial expression generation is one of the most challenging and long-sought aspects of character animation, with many interesting applications. The challenging task, traditionally having relied heavily on digital craftspersons, remains yet to be explored. In this paper, we introduce a generative framework for generating 3D facial expression sequences (i.e. 4D faces) that can be conditioned on different inputs to animate an arbitrary 3D face mesh. It is composed of two tasks: (1) Learning the generative model that is trained over a set of 3D landmark sequences, and (2) Generating 3D mesh sequences of an input facial mesh driven by the generated landmark sequences. The generative model is based on a Denoising Diffusion Probabilistic Model (DDPM), which has achieved remarkable success in generative tasks of other domains. While it can be trained unconditionally, its reverse process can still be conditioned by various condition signals. This allows us to efficiently develop several downstream tasks involving various conditional generation, by using expression labels, text, partial sequences, or simply a facial geometry. To obtain the full mesh deformation, we then develop a landmark-guided encoder-decoder to apply the geometrical deformation embedded in landmarks on a given facial mesh. Experiments show that our model has learned to generate realistic, quality expressions solely from the dataset of relatively small size, improving over the state-of-the-art methods. Videos and qualitative comparisons with other methods can be found at \url{https://github.com/ZOUKaifeng/4DFM}. Code and
models will be made available upon acceptance.
\end{abstract}



%
%
%
\maketitle

\section{Introduction}
\label{sec:intro}

3D facial expression synthesis is a fundamental, long-sought problem in face animation and recognition. Due to the inherent subtlety and sophistication of facial expressions, as well as our sensitivity to them, the task is extremely challenging yet highly beneficial to various multimedia applications, such as creating virtual humans in games and films, developing virtual avatars in immersive virtual and augmented reality experiences, implementing chatbots in digital marketing, and recognizing, managing, or tracking of emotional states for education and training purposes.
It has traditionally relied on time- and skill-intensive design work by trained CG (Computer Graphics) artists,  who utilized dedicated software equipped with geometric deformation and shape interpolation. The prevailing shape and motion capture technology has shifted this paradigm, allowing the algorithmic reconstruction of 3D face shape and motion of real people. Multi-view acquisition systems or 4D (3D+time) laser scans have been employed to capture the dynamic face geometry of the performer, even in real-time scenarios \cite{optideform1996, modelbased_tracking1999, facs2011, fperfcapture2011}. Such motion capture techniques have been supported by deformation transfer or animation retargeting \cite{rbface2000, reanimating2003, multilinear2006, Human2Avatar2020}, enabling the reuse of captured animations on new individuals or imaginary faces without having to capture shapes specific to the new target face.
This line of research has facilitated the construction of 4D face datasets and the derivation of priors from the data. Consequently, this advance has greatly amplified data-driven approaches to face modeling, paving the way for recent deep learning-based methods.   %
Reconstructive methods train the model to regress the 3D facial geometry from a 2D video, often in a frame-by-frame manner \cite{tu20203d, feng2018joint, deng2019accurate, guo2020towards} without modeling the temporal aspect of shape evolution. %
Generative models like Generative Adversarial Nets (GANs) \cite{gan2014} and Variational Autoencoders (VAEs) \cite{kingma2013auto} have been adapted to 
learn and sample the distribution of facial shapes and expression poses.
However, with a few exceptions \cite{G3AN, Seo_2021}, most existing works focus on the generation of 2D facial poses \cite{ganimation, fesgan2020}, or expression videos \cite{bouzid2022fev-gan,mocogan,imaginator, G3AN, naima_pami}, leaving the challenging task of 4D geometry modeling largely unexplored.

In this paper, we address the challenging problem of 3D dynamic facial expression generation, one that has not yet received a lot of attention. Most available 3D facial expression datasets \cite{zhang2013high, coma, cheng20184dfab, cosker2011facs, fanelli20103} come in the form of dense triangular meshes containing thousands of vertices. It is computationally expensive to train a generative model directly using all the vertices. Therefore, similarly to most successful models for 3D facial animation generation, we use a set of predefined 3D face landmarks to represent a face. Typically, landmarks are located on facial features that are highly mobile during animation, such as the face outline, eyes, nose, and mouth. The specific aim of the 3D facial animation generation is to learn a model that can generate facial expressions that are realistic, appearance-preserving, rich in diversity, with various ways to condition it such as categorical expression labels. Prior works that have attempted to model the temporal dimension of the face animation \cite{naima_pami, naima, Seo_2021, wang2018every} mostly leverage auto-regressive approaches, such as Long short-term memory (LSTM) \cite{lstm} and Gated recurrent units (GRUs) \cite{gru}. Here, we propose to use a Denoising Diffusion Probabilistic Model (DDPM) \cite{sohl2015deep, song2020score, ddpm}, a generative approach that has achieved remarkable success in several domains. %
A DDPM has the nice property of being trainable unconditionally whereas the reverse process can still be conditioned using, a classifier-guidance \cite{dhariwal2021diffusion}, for instance. 
This allows us to define the following paradigm: a DDPM is learned unconditionally and several downstream tasks associated with several conditional generations are developed from the same learned model, such as expression control (with label or text), expression filling (with partial sequence(s)), or geometry-adaptive generation (with facial geometry). This makes the proposed approach highly flexible and efficient, benefiting from the generative power of diffusion models while circumventing their limitations of being resource-hungry and difficult to control.

We note that, concurrent to this work, several works have also adopted diffusion models for human motion generation \cite{tevet2022human, zhang2022motiondiffuse, kim2022flame}. However, to the best of our knowledge, we are the first to adapt diffusion models to 3D face expression generation. More importantly, although approaches developed in \cite{tevet2022human, zhang2022motiondiffuse, kim2022flame} enable different forms of conditioning, they require the diffusion model to be retrained for each way of conditioning.

While the task of 3D facial animation generation has been reduced to the estimation of a temporal sequence of 3D face landmark sets, it is then necessary, in a second task, to compute a sequence of animated meshes. We use an  encoder-decoder model similar to \cite{naima}, which retargets the expression of a 3D face landmark set to the neutral 3D face mesh by computing its per-vertex displacement, in a frame-by-frame manner. %
Unlike \cite{naima}, however, we take into account the different morphological shapes of the neutral mesh to adapt the estimation of per-vertex displacements. Results thus obtained validate the effectiveness of the proposed approach. 

In summary, our key contributions are as follows: (1) We successfully use a DDPM to propose an original solution to the conditional generation of 3D facial animation. To the best of our knowledge, it is the first to adopt a diffusion-based generative framework in 4D face modeling. (2) We train a DDPM unconditionally and develop several downstream tasks by conditioning the reverse process. %
In addition to improving the efficiency of training, this paradigm makes the approach highly versatile and easily applicable to other downstream tasks.
(3) In various evaluations, the landmark sequence generation and landmark-guided mesh deformation outperform SOTA methods.

\section{Related work}
\label{related}

\subsection{Denoising Diffusion Probabilistic Models}
3D facial expression generation is achieved here by adapting a DDPM.
The surge in popularity of diffusion models is primarily due to this model \cite{ddpm}, which has made significant advancements in high-resolution image generation.

A Denoising Diffusion Probabilistic Model (DDPM) is a Markov chain aimed at synthesizing data (that resembles the training data) from noise within a finite number of sampling steps. The model's parameters undergo training via variational inference so as to reverse a diffusion process -- a Markov chain that gradually adds noise to the initial data.
Such diffusion models have a wide range of applications across various fields. In computer vision, diffusion models have demonstrated impressive capabilities for image generation\cite{latent, dalle2, imagen}, super-resolution \cite{li2022srdiff, saharia2022image, ho2022cascaded}, image inpainting \cite{inpainting}, image translation \cite{saharia2022palette, preechakul2022diffusion}, and semantic segmentation \cite{baranchuk2021label, brempong2022denoising, graikos2022diffusion}. They also demonstrated capability in handling 3D data to generate and complete point clouds \cite{zhou20213d, luo2021diffusion, lyu2021conditional}, as well as handling time series data for human motion generation \cite{tevet2022human, zhang2022motiondiffuse}, time series forecasting and imputation \cite{alcaraz2022diffusion}, and video generation \cite{harvey2022flexible, ho2022video, yang2022diffusion, ho2022imagen}. Additionally, diffusion models have demonstrated significant potential in the realm of natural language processing with numerous applications and use cases \cite{austin2021structured, li2022diffusion, chen2022analog, gong2022diffuseq, dieleman2022continuous}.
They find application in audio synthesis as well \cite{diffwave}.

However, diffusion models involve a complex generation process and suffer from slow sampling and have no encoding functionality. Several approaches have been proposed to address the limitations of DDPM. One such approach is Denoising Diffusion Implicit Model (DDIM) \cite{song2020denoising}, which accelerates the sampling process by reducing the number of required sampling steps. DDIM also allows for deterministic reverse processing, enabling various image editing possibilities \cite{hertz2022prompt}. However, DDIM inversion can result in instability and distorted reconstructions. To address this issue, \cite{wallace2022edict} proposes a novel approach inspired by the coupling layers in normalizing flow models \cite{dinh2014nice} providing mathematically exact inversion. In addition, several strategies exist to improve the performance of DDPMs, such as learning the variances of the reverse diffusion process, using a cosine noise schedule, and adding extra loss terms to optimize the variational lower-bound \cite{nichol2021improved}.

Note that each generative model has its own advantages and disadvantages. For example, VAE \cite{kingma2013auto} has nice encoding capabilities; however, it tends to lose high-frequency information of images. On the other hand, GANs \cite{gan2014} have the ability to produce high-quality images, but they are challenging to train and often prone to mode collapse. 
Therefore, each method has its own suitable application scenarios. For instance, diffusion models are preferred when prioritizing image quality over generation time. At the same time, efforts are underway to integrate different models aimed at overcoming their respective limitations, as seen in initiatives like DiffuseVAE \cite{pandey2022diffusevae} and Diffusion-GAN \cite{wang2022diffusion}.

\subsection{Conditional generation}
The conditional generation has been investigated since its inception, and its importance continues to increase. The generation can be conditioned by various modalities, ranging from a simple label to an image, or a sentence, which guides or controls the generation process. For example, image translation involves creating images conditioned on other images. Text-to-image generation uses text to control the characteristics of the generated images. In image captioning, conversely, the image conditions the text generation. %

The first approach to achieve conditional generation involves the use of conditional generative models, which are trained by incorporating specific conditions.
For instance, in the context of diffusion models, a commonly used approach is to train a label embedding and combine it with the time step embedding. The resulting combined information is then fed into the noise approximator, which generates the conditional noise samples \cite{li2022diffusion}.
Several models have been proposed to accommodate various forms of label conditions for GANs or VAEs, such as CVAE \cite{sohn2015learning}, CGAN \cite{mirza2014conditional}, PPGN \cite{nguyen2017plug}, and CVAE-GAN\cite{bao2017cvae}.  The neural network must include a mechanism to effectively incorporate the conditional information. One straightforward approach is to concatenate the label with an intermediate output of the model. In the context of conditional human motion generation \cite{actor}, learnable tokens are used. These tokens, of the same size as the latent representation, are added as offsets to the latent space.
We employ a comparable mechanism to integrate conditional information in our landmark-to-mesh deformation task (Sec. \ref{sec:lgmd}), focusing on generating a sequence of animated meshes from a sequence of landmarks, conditioned on the morphological shape of a neutral mesh.

Another line of approach related to ours is conditioning models that were originally learned unconditionally. This enables us to perform conditional generation using high-performance pretrained generative models. Note that the training data used for learning the ``independent conditioning task" is not required to be the same as the data used for training the generative model. Moreover, this paradigm enhances training efficiency, especially when there are multiple ways to apply conditioning. 
In this work, we unconditionally train a DDPM, after which we delve into several downstream tasks by conditioning the reverse process in different ways.
One commonly employed approach in conditioning generation within diffusion models involves classifier-guided sampling \cite{song2020score, dhariwal2021diffusion}. To condition a pre-trained diffusion model, it leverages the gradients of an independent classifier. This entails training a classifier on noisy data and subsequently using the gradients to guide the diffusion sampling process towards a specific class label.

Note that conditioning an unconditional model can also be achieved with other generative models than DDPM, especially through latent code manipulation \cite{shen2020interpreting,Collins20}. 
These methods begin by computing the latent representation of an image, and then adjusting this representation in a way that carries semantic meaning.

\subsection{Facial expression generation}

Deep generative models \cite{kingma2014semi, kingma2013auto, gan2014, flow, sohl2015deep} have proven effective at high-quality image synthesis, such as content-preserving image rendering with different styles, and the generation of images depicting learned objects. For 2D images, these models have also shown to be beneficial to facial expression transfer and expression editing tasks. However, the majority of existing solutions address the problem of \emph{static} expression synthesis. Here we review some recent advances achieved in \emph{dynamic} facial expression generation, i.e. modeling and predicting the temporal evolution of  poses elicited by facial expressions. \\

\noindent\textbf{2D facial expression video generation.} Building on the achievements in deep image generation, recent works have tacked the problem of generating 2D facial expression videos \cite{bouzid2022fev-gan,mocogan,imaginator, G3AN, naima_pami}. 
Methods like \cite{kim2018deepvideo} focus on facial expression transfer, where the dynamics of the facial expression is transferred from a source to a target face.
In other works, the dynamics of facial expression is encoded using recurrent generators. For example, Wang et al \cite{wang2018every} proposed to generate sequence of 2D landmarks by a conditional LSTM, which is trained over a manifold of landmark images found by a VAE. Addressing the one-to-many generation problem, they generate multiple sequences given a neutral face image, which are translated to face videos.
Recent methods seek to learn the spatiotemporal embedding and generate all frames simultaneously, rather than generating frames sequentially. 
G$^3$AN \cite{G3AN} learns disentangled representations of the facial video, appearance and motion by a three-steam generator, each taking into account appearance features, motion features, and the fusion of the two toward video generation.  
Other works have investigated modelling and learning the appearance and the motion in a disentangled manner. MoCoGAN \cite{mocogan} decomposes the video into content and motion: An image-based generator generates the content, an image representing the identity of the person, while a GRU-based motion GAN generates the motion-dependent appearance variations in videos through  sequential sampling from the motion space.
ImaGINator \cite{imaginator} presents an image to video generation consisting of an image encoder and a video decoder. The appearance extracted from the encoder is fused with motion feature in multi-level feature space via a spatio-temporal fusion in the video decoder to simultaneously generate appearance and motion, i.e. video. Two stream discriminator assesses the authenticity of the generated images and videos, respectively. Such an encoder-decoder based approach has been also adopted in FEV-GAN\cite{bouzid2022fev-gan}, where a an encoder is trained to extract identity- and spatial- features from the neutral face image, and a video decoder utilizes them along with a label to generate realistic video. 
\\

\noindent \textbf{Generative models for body animations.} Several works have shown the sequence generation on the body motion \cite{actor, tevet2022motionclip,li2021ai,li2022danceformer, guo2020action2motion}. This is mainly due to the compact, readily available skeleton-based representation of the body \cite{smpl}, the relatively large set of action vocabulary, and the availability of rich 3D body motion datasets \cite{mahmood2019amass, punnakkal2021babel, plappert2016kit, li2022danceformer, ionescu2013human3}. Unfortunately it is not yet the case with the 3D facial expression. 
\\

\noindent\textbf{Dynamic 3D facial expression synthesis.} %
To our knowledge, dynamic 3D facial expression synthesis has not been fully explored.
\cite{potamias2020} synthesizes realistic high resolution facial expressions by using a deep mesh encoder-decoder like architecture to estimate the displacements which are then added to a neutral face frame. \cite{Seo_2021} deploys LSTMs to estimate the facial landmark changes, which are then used to guide the deformation of a neutral mesh via a Radial Basis Function network. However, both works focus on the displacement estimation for a given expression and do not consider conditional generations.
The closest work to ours is Motion3DGAN \cite{naima} which extends the aforementioned MotionGAN \cite{naima_pami} to model the dynamics of 3D landmarks. The learned distribution of 3D expression dynamics by a WGAN over the hypersphere space is sampled with a condition to generate landmark sequences, which are then fed into a mesh decoder to deform a neutral 3D face mesh frame-by-frame. Our work has several advantages over their work. %
Benefiting from the power of diffusion models, we model the input distribution  without requiring any extra preprocessing, and can learn from sequences of different lengths. 
Moreover, our framework offers a highly versatile and efficient alternative, as we train a DDPM unconditionally and different conditional generations can be performed solely during the reverse process in a plug-and-play manner. %

Given the scarcity of existing work on 3D facial animation generation, we compare our work with some generator models originally dedicated to human motion synthesis, including Action2motion \cite{guo2020action2motion} and ACTOR \cite{actor}.

\section{Method}
\label{sec:med}
\begin{figure*}[t]
\begin{center}
   \includegraphics[width=1\linewidth]{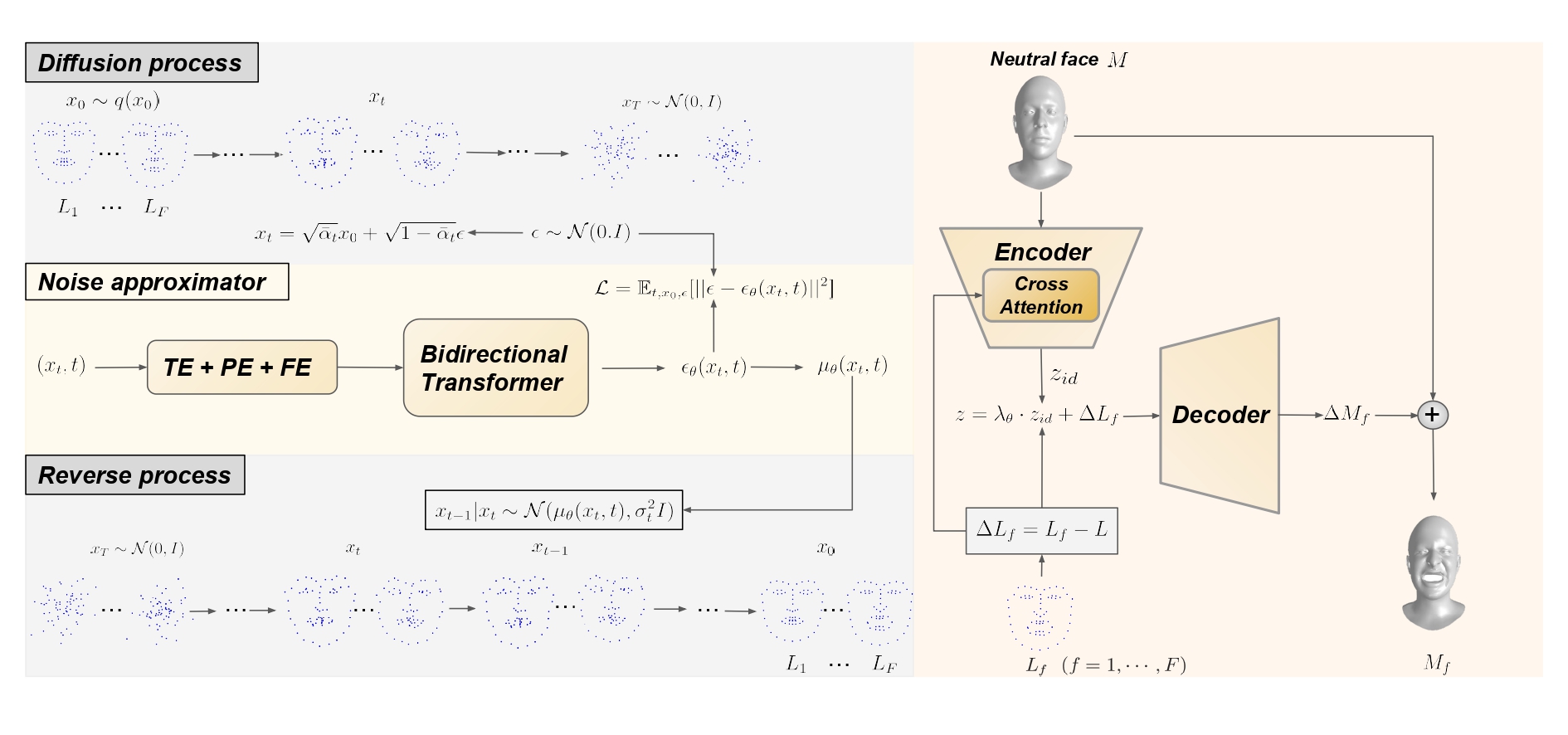}
\end{center}
   \caption{Overview of the proposed approach. Generally, the diffusion process is used to train the noise approximator while the reverse process is used to sample $x_0$ from the distribution $q$. But some tasks developed in Sec. \ref{sec:downstrm} require both processes for sampling. The bidirectional transformer takes as input the sum of the outputs of three embedding layers: the temporal embedding layer (TE) that takes as input $t$, the positional encoding layer (PE) that takes as input an integer sequence from $1$ to $F$, and the feature embedding layer (FE) that takes $x_t$. The landmark-guided encoder-decoder retargets the expression of $L_f$ onto the input mesh $M$ to estimate $M_f$ at each frame.}
\label{fig:model}
\setlength{\abovecaptionskip}{0.cm}
\end{figure*}

At the core of our approach is a DDPM-based model to generate a 3D landmark sequence $x=\{L_1,\ldots,L_F\}$ where a frame $L_f\in \mathbb{R}^{N\times3}$ (for $f=$1 to $F$) represents the 3D coordinates of $N$ landmarks.
Note that the 3D arrangement of a landmark set $L_f$ implicitly encodes the geometric information specific to the facial anatomy of an individual, and can be viewed as a mixture of the facial identity shape at a neutral pose $L$ and the pose-induced shape change, i.e. $L_f=\Delta L_f+L$. The method is composed of two tasks: First, a DDPM is trained unconditionally (Sec.\ref{sec:ddpm}), whereas conditional generations are obtained by conditioning the reverse process. Different forms of conditioning can be performed, leading to several downstream tasks (Sec.\ref{sec:downstrm}). Afterwards, our landmark-guided encoder-decoder (Sec.\ref{sec:lgmd}) estimates $\Delta M_f$ at each frame  (for $f=$1 to $F$), using a target neutral face mesh $M$ and $\Delta L_f$ as input. The desired animation mesh sequence $\{M_1,\ldots,M_F\}$ is obtained by adding the estimated displacement $\Delta M_f$ to $M$ at each corresponding frame, i.e. $M_f=M + \Delta M_f$. The overview of the proposed method is illustrated in Fig.\ref{fig:model}.

Note that directly training from and generating full meshes may be beneficial but raises technical issues since the model becomes computationally and memory intensive. An alternative is to utilize diffusion models directly in the latent space of autoencoders \cite{latent}, or a pre-constructed parameter space of 3D face. Our work can be viewed as akin to the latter approach, except that we use a heuristically defined feature space, i.e., the landmark space,
instead of 
a learned latent space. This choice has been validated by the quality of the reconstruction obtained by the landmark-guided encoder-decoder (Tab. \ref{tab:recon}).

\subsection{Denoising Diffusion Probabilistic Models}
\label{sec:ddpm}
DDPMs are latent variable models where the latent variables $x_t$ (for $t=$1 to T) have the same dimension as the original data $x_0 \sim q(x_0)$. In our work, $x_0$ is a landmark-based facial animation data: $x_0 = \{L_1,\ldots,L_F\}$. Note that it is contrary to most prior works which generate only the displacements $\Delta L_f$ \cite{naima_pami, naima, Seo_2021}. Training our model to generate $L_f$ directly allows it to learn to produce quality expressions that are consistent with the inherent facial morphology. 

DDPM consists of two distinct processes. The unparameterized diffusion process is a Markov chain, that gradually adds noise to the initial data. Conversely, the parameterized reverse process involves a Markov chain specifically designed to generate, within a limited number of sampling iterations, data resembling the training dataset and originating from noise. The parameters of the reverse process are trained so as to reverse the diffusion process.

More formally, the joint distribution $p_{\theta}(x_{0:T})$ from which we derive the likelihood $p_{\theta}(x_0) = \int p_{\theta}(x_{0:T}) dx_{1:T}$ is the reverse process whereas the approximate posterior $q(x_{1:T} |x_0)$ is  the forward process or diffusion process. The diffusion process produces gradually noisier samples ($x_1, x_2, \ldots x_T$) by adding Gaussian noise to the initial data $x_0$ according to a variance schedule $\beta_1,...,\beta_T$\cite{ddpm}:
\begin{eqnarray}
q(x_{1:T}|x_0) &=& \prod_{t=1}^T q(x_{t}|x_{t-1}) \\
    q(x_{t}|x_{t-1}) &=& \mathcal{N}(x_t; \sqrt{1-\beta_t}x_{t-1}, \beta_t I).
\label{eq:dd}
\end{eqnarray}
We can derive from Eq. \ref{eq:dd} the following property \cite{ddpm} which allows us to train the diffusion model efficiently at an arbitrary time step $t$:
\begin{equation}
q(x_t|x_0) = \mathcal{N}(x_t; \sqrt{\bar{\alpha}_t}x_0, (1-\bar{\alpha}_t)I),
\label{eq:qt}
\end{equation}
where $\bar{\alpha}_t = \prod^t_{s=1} \alpha_s$ and $\alpha_t = 1-\beta_t$. 

$x_T$ follows a near-isotropic Gaussian distribution provided that a well-behaved schedule is defined and that $T$ is sufficiently large.
DDPM \cite{ddpm} uses this property to sample the target distribution $q$ ($x_0 \sim q(x_0)$). This is achieved by reversing the diffusion process: It begins by sampling $x_T$ from $\mathcal{N}(0,I)$. Next, the reverse process generates progressively less-noisy samples $x_{T-1}, x_{T-2}, \ldots, x_1$ until $x_0 \sim q(x_0)$ is obtained, by repeatedly sampling $x_{t-1}$ from $p_{\theta}(x_{t-1}|x_t)$ by using Eq. \ref{eq:lm}.
This reverse process is formally defined as a Markov chain with learned Gaussian transitions whose mean and variance are estimated by a neural network of parameter $\theta$:
\begin{eqnarray}
    p_{\theta}(x_{0:T}) &=& p(x_T) \prod_{t=1}^T p_{\theta}(x_{t-1}|x_t) \\
    p_{\theta}(x_{t-1}|x_t) &=&
    \mathcal{N}(x_{t-1}; \mu_\theta(x_t,t), \Sigma_\theta(x_t,t)), \
    \label{eq:lm}
\end{eqnarray}
where $p(x_T) = \mathcal{N}(x_T;0,I)$. As in \cite{ddpm}, we set $\Sigma_\theta(x_t,t)$ to $\sigma^2_t I$. This is a reasonable choice for generating quality samples, provided that $T$ is chosen to be sufficiently large \cite{nichol2021improved}. 
Note that estimating $\Sigma_\theta(x_t,t)$ allows sampling with many fewer steps \cite{nichol2021improved}. 

Several possibilities can be considered to parameterize $\mu_\theta(x_t,t)$ in Eq. \ref{eq:lm}. \cite{ddpm} shows that approximating the noise $\epsilon$ that appears in the following equation:
\begin{equation}
x_{t} = \sqrt{\bar{\alpha}_t}x_0 +   \sqrt{1-\bar{\alpha}_t} \epsilon, 
\label{eq:cop}
\end{equation}
is a suitable choice, especially when combined with a simple loss function (See Eq. \ref{eq:lossf}). Note that Eq. \ref{eq:cop} is a different way of writing Eq. \ref{eq:qt} ($\epsilon \sim \mathcal{N}(0,I)$). Finally, the term $\mu_{\theta}(x_t,t)$ can be computed from the approximation of $\epsilon$, denoted as $\epsilon_\theta(x_t,t)$:
\begin{equation}
\mu_{\theta}(x_t,t)= \frac{1}{\sqrt{\alpha_t}}\left(x_t-\frac{\beta_t}{\sqrt{1-\bar{\alpha}_t}} {\epsilon}_\theta\left(x_t, t\right)\right).
\end{equation}

Diffusion models can be trained by optimizing the usual variational bound on negative log-likelihood, but we adopt here the simplified objective function proposed in \cite{ddpm}:

\begin{equation}
 \mathbb{E}_{t,x_0, \epsilon}[||\epsilon - \epsilon_\theta(x_t, t) ||^2],
\label{eq:lossf}
\end{equation}
where the term $x_t$ is computed from Eq. \ref{eq:cop}.

Many previous works \cite{ddpm,nichol2021improved, latent, dhariwal2021diffusion}, especially those for modeling 2D images, have utilized a UNet-like structure\cite{unet} to model the mean $\mu_{\theta}(x,t)$ or the noise $\epsilon_{\theta}(x,t)$. Here we employ 
a bidirectional transformer (BiT) \cite{bert} to efficiently capture the temporal characteristics of $x_t$.

\subsection{Downstream tasks}
\label{sec:downstrm}
The DDPM is trained in an unconditional manner, and from this single trained model, various downstream tasks can be derived. 
Here, we demonstrate 
expression control tasks through labels or texts, and expression filling using partial sequence, offering tools for creators to explore new expression sequences. 
The last downstream task, geometry-adaptive generation,
enables us to generate a landmark sequence conditioned to a  facial anatomy.%
The pseudo code for each task can be found in Appendix \ref{Pseudocode}.

\noindent \textbf{Conditioning on expression label (label control).}
The task is to perform a conditional generation according to the expression label $y$. Conditioning the reverse process of an unconditional DDPM is achieved by using the classifier-guidance \cite{song2020score, dhariwal2021diffusion, li2022diffusion}. 
First, we train a classifier that predicts the label $y$ given a latent variable $x_t$ (and $t$). %
Here the classification is conducted with a BiT by adopting the usual approach of adding an extra learnable classification token \cite{bert}. 
Note that the BiT discussed here should be distinguished from the other BiT in the diffusion model and is used to condition its reverse process. 
It is achieved by sampling $x_t$ according to the distribution:
\begin{equation}
p_{\theta,\phi}(x_{t}|x_{t+1}, y) \propto p_{\theta}(x_t|x_{t+1}) p_{\phi}(y|x_t), 
\label{eq:coucou}
\end{equation}
where $\phi$ represents the parameters of the classifier. Sampling of Eq. \ref{eq:coucou} can be achieved approximately \cite{sohl2015deep} by sampling from a Gaussian distribution similar to the unconditional transition operator $p_{\theta}(x_t|x_{t+1})$, but with its mean shifted by a quantity proportional to $\Sigma_{\theta}(x_t,t) \nabla_{x_t}p_\phi(y|x_t)$.  

Instead of sampling Eq. \ref{eq:coucou}, we used an alternative way, as proposed in \cite{li2022diffusion}: $x_t$ is computed so as to maximize the $log$ of Eq. \ref{eq:coucou}. A hyperparameter $\lambda$ is used to adjust the trade-off between fluency ($p_{\theta}(x_t|x_{t+1})$) and control ($p_{\phi}(y|x_t)$), leading to a stochastic decoding method that balances maximizing and sampling $p_{\theta,\phi}(x_t| x_{t+1}, y)$. As in \cite{li2022diffusion}, optimization is achieved by running 3 steps of the Adagrad \cite{duchi2011adaptive} update for each diffusion step (Alg. \ref{algo:label} of App. \ref{Pseudocode}).

\noindent \textbf{Conditioning on text (text control).}
We also use in this task a BiT guidance, but instead of estimating a label from $x_t$ and $t$, the BiT outputs a vector of dimension 512 (the softmax layer is removed). As in \cite{tevet2022motionclip}, the BiT is trained so as to increase the cosine similarity between its output  and the textual features extracted with CLIP \cite{clip} from the text associated with $x_0$.
Conditioning the reverse process according to the text $c$ is then achieved (Alg. \ref{algo:text} of App. \ref{Pseudocode}) by adapting the procedure presented for the label control: $x_t$ is computed so that it maximizes:
\begin{equation}
\lambda\cdot log(p_\theta(x_t|x_{t+1}))+ cos(\mbox{BiT}(x_t,t), \mbox{CLIP}(c)).
\end{equation}

\noindent \textbf{Conditioning on partial sequence (expression filling).} Similarly to inpainting whose purpose is to predict missing pixels of an image using a mask region as a condition, this task aims to predict missing frames of a temporal sequence by leveraging known frames as a condition. The sequence $x_0$ is composed of $F$ frames, some of which are unknown. Let $\mathcal{S}_K$ and $\mathcal{S}_U$ denote respectively the set of indices associated with known and unknown frames, and let $x|_{\mathcal{S}}$ denote the subsequence containing only the frames of $x$ whose indices belong to $\mathcal{S}$.

Since ${x_0}|_{\mathcal{S}_K}$ is known, note that ${x_t}|_{\mathcal{S}_K}$ can be  drawn according to Eq. \ref{eq:qt}. Indeed, each component of $x_t$ can be drawn independently since $q(x_t|x_0)$ is an isotropic normal distribution. %
Sampling from the reverse process conditioned on a partial sequence can also be achieved as follows: $X_T$ is first determined: ${x_T}|_{\mathcal{S}_U}$ is drawn from $\mathcal{N}(0,I)$ and ${x_T}|_{\mathcal{S}_K}$ according to Eq. \ref{eq:qt}. Then, computing $x_{t}$ from $x_{t+1}$ is achieved in two steps: First, a temporal sequence $\hat{x}_{t}$ is simply drawn from $p_{\theta}(.|x_{t+1})$ (it is the way to compute $x_{t}$ in the usual case). ${x_{t}}|_{\mathcal{S}_U}$ is set to $\hat{x}_{t}|_{\mathcal{S}_U}$, while for known frames, ${x_{t}}|_{\mathcal{S}_K}$ is directly drawn according to Eq. \ref{eq:qt} (Alg. \ref{algo:filling} in App. \ref{Pseudocode}).
Despite its simplicity, this strategy gives satisfactory results as we will demonstrate through qualitative validation in later sections of this paper, provided that the partial sequence is of sufficient length.

\noindent \textbf{Geometry-adaptive generation.}
Given the facial geometry of a specific subject, a generation can be performed as a special case of expression filling: 
${\mathcal{S}_K}$ is set to $\{1\}$ or to $\{F\}$ ($F$ is the sequence length) and the unique known frame associated with ${x_{0}}|_{\mathcal{S}_K}$ is set to the neutral face $L$ of the subject. 
The remaining sequence is considered as unknown, for which the model performs an expression filling.

However, we observed that the generated frames may not always smoothly connect to the given frame, a problem that did not arise when the partial sequence remained long enough. %
In the context of image inpainting, \cite{inpainting} also shows that the simple sampling strategy used for the expression filling task may introduce disharmony. A more sophisticated approach has been proposed so as to harmonize the conditional data ${x_{t}}|_{\mathcal{S}_K}$ with the generated one ${x_{t}}|_{\mathcal{S}_U}$ \cite{inpainting}. 
In order to achieve better convergence properties of the algorithm while maintaining its simplicity,  we derive the sequence with five iterations, each with a slight modification: For the first iteration, ${x_T}|_{\mathcal{S}_U}$ is drawn, as previously, from $\mathcal{N}(0,I)$. For the following iterations, ${x_T}|_{\mathcal{S}_U}$, as ${x_T}|_{\mathcal{S}_K}$, is drawn according to Eq. \ref{eq:qt} where $x_0$ is the result obtained from the previous iteration.
By doing so, we expect ${x_T}|_{\mathcal{S}_U}$ and ${x_T}|_{\mathcal{S}_K}$ to be harmonized progressively, thus leading to the improved harmonization of ${x_t}|_{\mathcal{S}_U}$ and ${x_t}|_{\mathcal{S}_K}$ along the iterations. 

Note that this process can also be easily guided by a classifier (as in the label control) so as to generate a desired facial expression starting from a given facial anatomy (See Alg. \ref{algo:geo} in App. \ref{Pseudocode}). %

\subsection{Landmark-guided mesh deformation}
\label{sec:lgmd}
To obtain the full mesh sequence $\{M_1,\ldots,M_F\}$ from $\{L_1,\ldots,L_F\}$, one could use existing fitting methods such as FLAME \cite{flame} or DL-3DMM \cite{ferrari2017dictionary} so as to preserve both the facial anatomy and the expression encoded in the landmark frames. However, the meshes generated through the linear blending models tend to lack intricate details of facial geometry, resulting in dull, lifeless shapes.
Thus, in our work, we retarget the expression encoded in $L_f$ to the facial geometry given as a (realistic) input mesh $M$, as in \cite{naima}. 
The mesh $M$ is assumed to be at its neutral pose with a predefined topology \cite{flame}. 
Each mesh frame $M_f$ should retain the facial identity shape $M$, combined with the expression-driven shape change encoded in $\Delta L_f=L_f-L$ ($\Delta L_f$ represents landmark displacements at $f$-th frame). This is achieved by our encoder-decoder network that takes both $M$ and $\Delta L_f$ as input and 
predicts $\Delta M_f$ at each frame, which is respectively added to $M$ to obtain the final mesh sequence: $M_f=M +\Delta M_f$. 
This is similar  to the Sparse2Dense mesh decoder proposed in \cite{naima}, except that only $\Delta L_f$ (and not $M$) is used to predict $\Delta M_f$ in their work.
In our approach, on the other hand, we take into account the different morphological shapes of the neutral mesh $M$ to adapt the estimation of per-vertex displacements $\Delta M_f$.

In order to benefit from the 
consistent and quality expressions 
adapted to the facial morphology by the DDPM, one can extract a landmark set $L_M$ from a mesh $M$, perform the geometry-adaptive task on it to generate a sequence involving $L_M$, and target it to $M$ by the landmark-guided mesh deformation.

\noindent \textbf{Encoder \& decoder.} Inspired by the Sparse2Dense mesh decoder
of \cite{naima}, we develop an encoder-decoder architecture based on spiral operation layers.
The encoder contains a backbone consisting of five spiral operation layers \cite{spiral} that extracts the features of $M$. %
In addition, we propose to incorporate a cross-attention mechanism \cite{transformer} to account for the possible influence of the characteristics of $M$ on the impact of $\Delta L_f$ on each vertex of $M$:
It enables us to find the relevant features of the mesh $M$ that can help predict a latent representation (of $\Delta M_f$) according to $\Delta L_f$. 
More specifically, 
the \emph{query} is derived from a linear embedding of $\Delta L_f$ (computed by a fully-connected layer $FC$) and the \emph{key, value} pairs from the output of the backbone (i.e. features of $M$) denoted as $Fe$. 
The output of the attention layer writes:
\begin{equation}
softmax \left(\frac{ FC(\Delta L_f) \cdot F_e^T}{\sqrt{d}} \right)F_e,
\label{eq:GeDn2}
\end{equation}
where $d$ is the dimension of $F_e$.
Then a linear layer maps the vector of Eq. \ref{eq:GeDn2} to the \emph{identity-aware} representation $z_{id}$, which is further shifted by the landmark displacement $\Delta L_f$ to obtain the final latent representation: $z = \lambda_\theta \cdot z_{id} + \Delta L_f$, where the weight parameter $\lambda_\theta$ is a learnable parameter. 

We use the same decoder as \cite{naima}. It consists of a linear layer and five spiral operation layers. It takes the latent representation $z$ as input and outputs the per-vertex displacement $\Delta M_f$. $M_f$ is then set to $M + \Delta M_f$. 
The model is learned using the loss function proposed in \cite{naima}.

\section{Experimental setting}
\label{sec:dataset}
As proposed in \cite{ddpm}, we set a linear noise schedule starting from $\beta_1=1e-4$ to $\beta_T=0.02$, and $\sigma_t^2$ is set to $\beta_t$. $T$ is set to 2000. We train the model on $200$K iterations with a learning rate of $1e-4$ and a batch size of $256$. The hyperparameter $\lambda$ that is used to guide the sampling of the reverse process is set to $0.01$ as in \cite{li2022diffusion}.

\textbf{CoMA dataset} \cite{coma} is a commonly used 4D facial expression dataset in face modeling tasks \cite{spiral, jiang2019disentangled}, consisting of over a hundred 3D facial animation sequences captured from $12$ subjects, each performing $12$ facial actions (``high smile'', ``mouth up'', etc.). Each data is composed of a triangular mesh of $5023$ vertices undergoing some deformation elicited by an expression.%

\textbf{BU-4DFE dataset} \cite{zhang2013high} contains a total of $606$ sequences of $83$ landmarks extracted from a sequence of 3D facial scans. Six basic emotional expressions (``anger'', ``disgust'', ``fear'', ``happy'', ``sad'', and ``surprise'') of 101 subjects have been recorded. %

To efficiently capture the essence of expressions while maintaining a small data size, we use $68$ landmarks defined by FLAME\cite{flame} for both databases. The different sequences have been manually divided into sequences of around $40$ frames. Some of them start from the neutral pose and end with a maximal expression intensity (we call them sequences of type N2E), while others evolve from the maximal expression intensity to the neutral face (they are called of type E2N). Since a majority of the methods utilized for comparison necessitate the sequences to share a same length, linear interpolation has been carried out so as to obtain sequences of length 40. Note, however, that our method can deal with sequences of various lengths (See App. \ref{length}). 
As a result, our dataset comprises $689$ sequences from the CoMA dataset and $1,212$ sequences from the BU-4DFE dataset, totaling $1,901$ sequences with 18 facial actions. 
Different sequences have been used depending on the specific task at hand.
Unless otherwise specified, solely the sequences from the CoMA dataset have been utilized.

\section{Results involving various conditional generations}
Here we describe the results we obtained on the various conditional generations. Throughout this section, a classifier that predicts the expression from a sequence independently of its type (see Section \ref{sec:dataset}) is called a classifier of type I (order-\textbf{I}nsensitive), whereas a classifier of type S (order-\textbf{S}ensitive) predicts both the expression class and the expression type (either N2E or E2N).

To assess the fidelity of the generated expressions in relation to the control,
an independent classifier (which we denote as IC) is trained to predict the label from a sequence $x_0$. We use one LSTM layer followed by a linear layer, as in \cite{naima}. The model's ability to generate a desired expression is assessed by the classification accuracy of the IC tested on the generated expressions. Additionally, the quality and the diversity of the generated sequences are assessed by using the Frechet Inception Distance (FID) score \cite{fid}, 
a widely used metric for comparing the distribution of fake data with that of real data. It is computed from the output of the linear layer of the IC.

\subsection{Label control}
\label{sub:sec:labelcontrol}

\begin{figure}[ht!]
\centering
\includegraphics[width=0.71\linewidth]{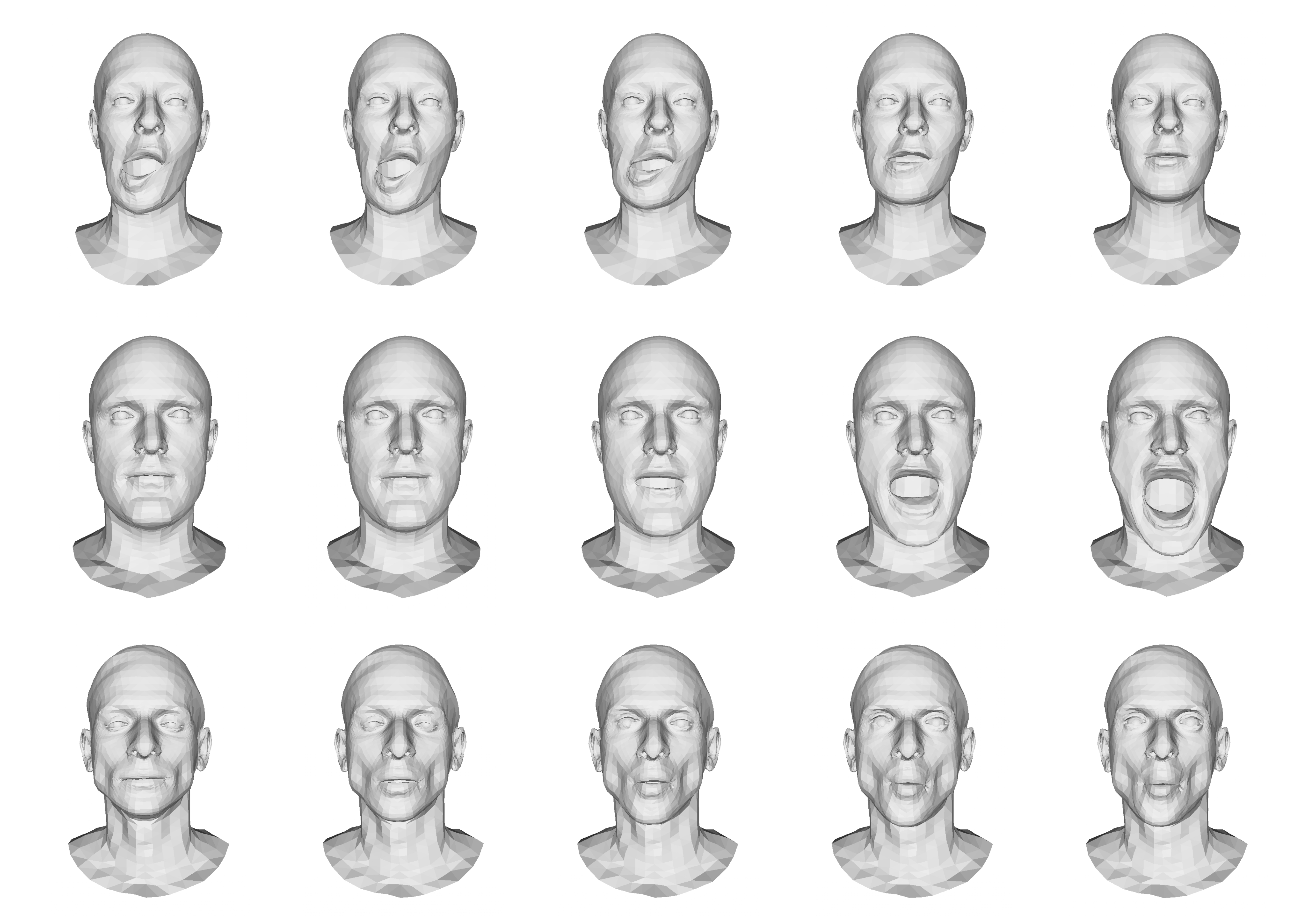}
\caption{Animated mesh sequences guided by the label ``mouth side'' (top),  ``mouth extreme'' (middle), and ``cheeks in'' (bottom). The meshes are obtained by retargeting the expression of the generated $x_0$ on different neutral faces.} %
\label{fig:conGen}
\end{figure}

The proposed approach is compared with several SOTA methods that perform conditional sequence generation: Action2Motion \cite{guo2020action2motion}, Motion3DGAN \cite{naima} and ACTOR \cite{actor}. The BiT-based classifier is used to guide the reverse process, as well as the IC are of type I. Quantitative evaluation results as measured by the classification accuracy and the FID score are summarized in Table \ref{tab:con}, which confirms that the proposed approach outperforms all SOTA methods. %
Fig. \ref{fig:conGen} shows some illustrative results: Our model generates various realistic and quality expressions adapted to different facial geometries.
Videos presented in the project website (\url{https://github.com/ZOUKaifeng/4DFM}) demonstrate the generated expressions and offer qualitative comparisons among these methods. Visual observation of these videos confirms that sequences generated by our approach are more expressive. 
The diversity of the generated sequences in terms of both expression and facial anatomy is illustrated in Appendix \ref{sec:diversity}.

 \begin{table}[ht!]

\caption{Performance of different methods for generating desired expressions has been evaluated by measuring the classification accuracy and the FID score. 
We report as ground truth the FID and the accuracy computed on the test dataset, assuming that an ideal method could have generated it.}
\centering

\begin{tabular}{lllll}
\hline\noalign{\smallskip}
      & \multicolumn{2}{c}{CoMA} & \multicolumn{2}{c}{BU-4DFE}  \\
  \cmidrule(r){2-3}  \cmidrule(r){4-5}
Model & Acc   & FID & Acc  & FID \\
\hline
Ground truth & $83.78\% $  & $2.77 $  &  $99.51\%   $ & $6.02  $ \\
A2M \cite{guo2020action2motion}& $52.36\% $  & $29.44$  &  $80.83\%   $ & $19.64 $ \\

MoGAN \cite{naima} & $80.76\% $ & $7.72$  &  $99.26\%$ & $13.29$\\
ACTOR \cite{actor}& $81.40\%  $   & $7.11 $ &  $99.13\%$ & $14.56 $ \\
Ours&$84.97\% $ & $6.79$ & $99.89\%$ & $12.37$ \\
\noalign{\smallskip}

\hline
\label{tab:con}
\end{tabular}
\end{table}

\subsection{Text control}
To demonstrate this task, we have increased the vocabulary of our dataset by merging CoMA and BU-4DFE. In the first experiment, the raw text label is used to condition the animation (we call it \emph{raw text} task) and the IC used for the evaluation is of type I. In the second experiment, the description of a sequence is enriched to be a short sentence such as ``from the neutral face to the raw text label'', or ``from the raw text label to the neutral face'' (we call it \emph{enriched text} task) 
and the IC used for the evaluation is of type S. 

We compare our results with those of MotionClip \cite{tevet2022motionclip}. 
Quantitative results are shown in Table \ref{tab:text_con}. 
Classification accuracies obtained with the proposed method are slightly higher than those of MotionClip, with FID scores significantly lower. Sequences created by MotionClip are actually realistic but the FID scores are high, due to the lack of diversity in the generated sequences. 

\begin{figure}[ht!]
\centering

\includegraphics[width=0.71\linewidth]{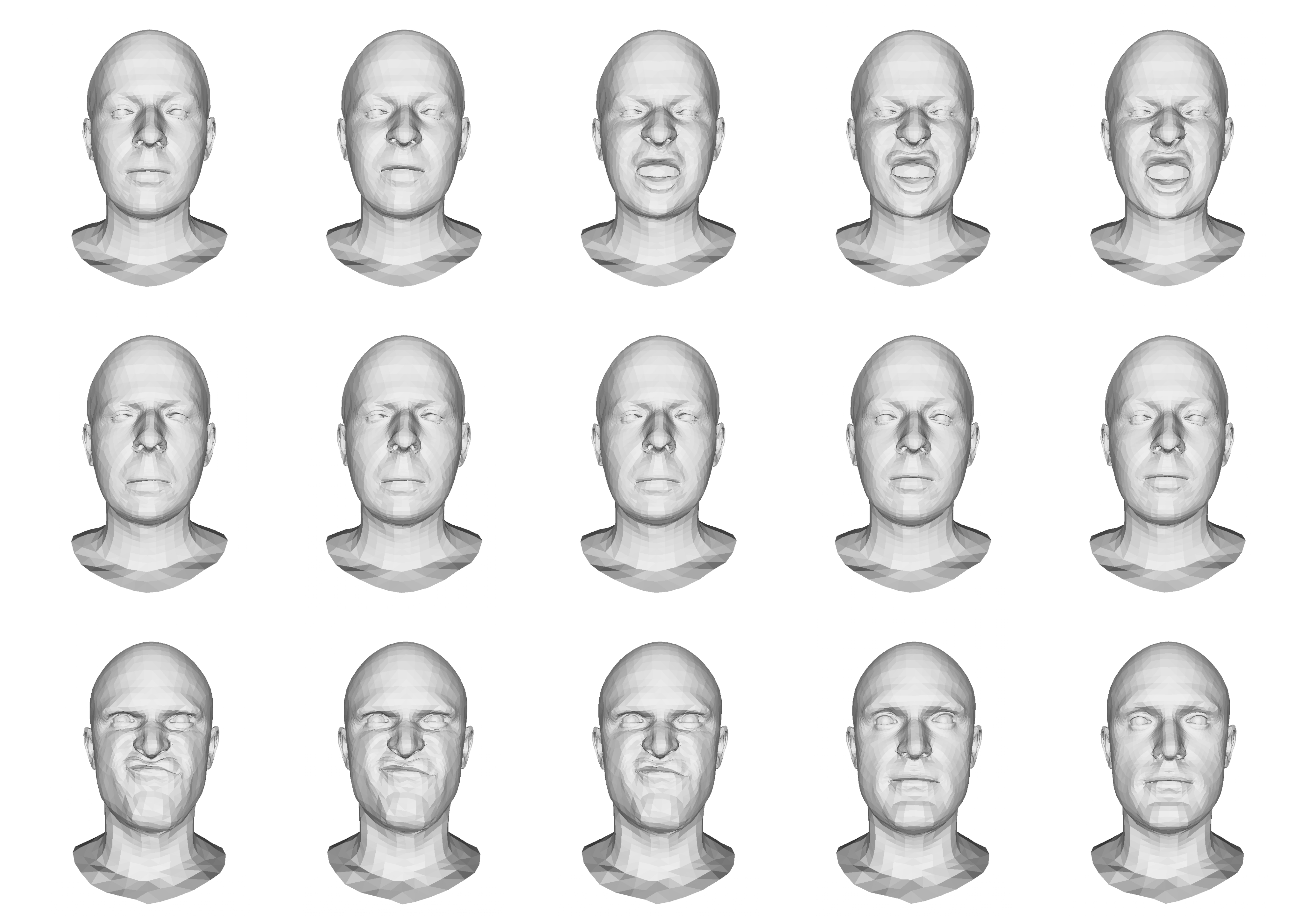}

\caption{Text-driven generation results obtained by the \emph{enriched text} task (``from neutral face to bareteeth'' (top)), and by the \emph{raw text} task (``angry mouth down'' (middle), ``disgust high smile'' (bottom)). The input texts used for the \emph{raw text} task are the combinations of two terms used for training. For instance, ``disgust high smile'' is a new description that hasn't been seen before, which combines ``disgust'' and ``high smile''. 
}
\label{fig:textGen}
\end{figure}

Fig. \ref{fig:textGen}
shows illustrative examples obtained  with the proposed approach. Note that our model is able to create animated meshes that combine different types of expressions by compositing a text combining different types of expressions.
For the complete sequences as well as the qualitative comparisons, readers may refer to the project website.

 \begin{table}[ht!]

\caption{Quantitative evaluation of the text control task. 
Classification accuracy and FID are computed for the raw text task (rtt, left) and for the enriched text task (ent, right).}
\centering

\begin{tabular}{lllll}
\hline\noalign{\smallskip}
     & Acc (rtt) & FID  & Acc (ent) & FID  \\
\hline
Ground truth & $86.02\%$& $3.67$ & $74.40\%$ & $4.56$ \\ 
MotionClip &$80.67\%$  &$42.19$ &$ 58.33\%  $ & $38.83$\\
Ours &$82.01\%$ & $9.46$ & $64.38\%$ & $11.34$\\ %
\noalign{\smallskip}
\hline
\label{tab:text_con}
\end{tabular}
\end{table}

\subsection{Expression  filling}

Given a partial sequence of an expression, the model can fill up the missing frames. %
Three experiments have been conducted: %
In the filling from the beginning (FFB) or the filling from the end (FFE) cases, the length $l$ of the partial sequence is drawn uniformly in $[10,30]$.
In the filling from the middle (FFM) case, $l$ frames have been given at the beginning and at the end of the sequence, respectively. $l$ is uniformly sampled in $[5,15]$.

The proposed approach for expression sequence filling is compared with a mean imputation strategy. To evaluate the result, an IC (of type I) is trained, so as to check if the filled data has the same expression class as the original one.
Results are shown in Table \ref{tab:filling}. 
The expression label of the partial sequence is well-captured and reflected in the filled part, leading to an improved classification accuracy especially for the FFM case, where the classification accuracy is comparable to that obtained for the ground truth (Table \ref{tab:con}). Classification accuracies obtained in the FFE and FFB cases are lower due to the content of the sequences. As an example, when the partial sequence is associated with the beginning of a sequence of type N2E, it may be composed, at worst, of neutral faces only, or at best of less expressive faces. This is worsened by the fact that sometimes certain expressions appear only at the end of the sequences. This is contrary to the FFM case, where the partial sequence contains both the neutral and the most expressive poses.

Finally, there is a significant improvement of FID score after filling with the proposed approach. Furthermore, our videos presented on the project website illustrate that the generated sequences are smoothly connected to the given partial sequence.

 \begin{table}[ht!]
\caption{Quantitative evaluation of the expression filling task for three different locations of the missing part. Accuracy and FID are computed on the sequences obtained by the mean imputation strategy, and by our diffusion model. Note that accuracy is 83.78\% and FID is 2.77 for the ground truth in all cases (FFE, FFM, FFB).}
\centering

\begin{tabular}{lllll}
\hline
\noalign{\smallskip}
      & \multicolumn{2}{c}{Mean Imputation} & \multicolumn{2}{c}{Ours}  \\
 \cmidrule(r){2-3}  \cmidrule(r){4-5}
 &   Acc &  FID  & Acc  &  FID  \\
\hline
FFE    &$60.15\%$  & $25.67$ & $67.18\%$ & $5.51$ \\ 

FFM   &$56.25\%$  & $17.68$ & $85.93\%$ & $5.06$ \\ 

FFB  &$53.90\%$  & $27.32$ & $70.31\%$ & $5.22$ \\

\noalign{}
\hline
\label{tab:filling}
\end{tabular}
\end{table}

\subsection{Geometry-adaptive generation}
We have conducted the  geometry-adaptive generation task 
by using classifier guidance so as to generate a desired facial expression from a given facial anatomy (Alg. \ref{algo:geo} of App. \ref{Pseudocode}).
The BiT used for guidance and the IC used for evaluation are both of type S. ${\mathcal{S}_K}$ is set to $\{1\}$ if the chosen label is associated with N2E sequences, and to $\{F\}$ otherwise.

Quantitative results are shown in Table \ref{tab:final}. 
The classification accuracy is close to the ground truth, and the visual inspection of the video sequences on the project website shows no gap between the generated frames and the enforced one. %

\begin{table}[]
\caption{Quantitative evaluation of the geometry-adaptive generation task.}
    \centering

\begin{tabular}{llll}
\hline\noalign{\smallskip}
     & Acc & FID  \\
\hline
Ground truth & $71.01\%$ & $5.57$ \\
Geometry-adaptive &$ 70.43\% $ & $9.26$  \\
\noalign{\smallskip}
\hline
\end{tabular}
\label{tab:final}
\end{table}

While App. \ref{sec:diversity} illustrates the diversity of generated expressions when the model is conditioned on the expression label, we study here the same type of diversity but when the facial geometry of a specific subject is enforced in the conditioning process. To this end, a landmark set $L_M$ has been extracted from a given mesh $M$. The geometry-adaptive generation task is performed so as to generate a sequence containing $L_M$, and exhibiting an expression corresponding to a given label $y$. Then, the generated sequence is targeted to $M$ with the landmark-guided mesh deformation. 

\begin{figure}[h]
  \centering
    \begin{minipage}{.45\textwidth}
       
  \includegraphics[width=1\linewidth]{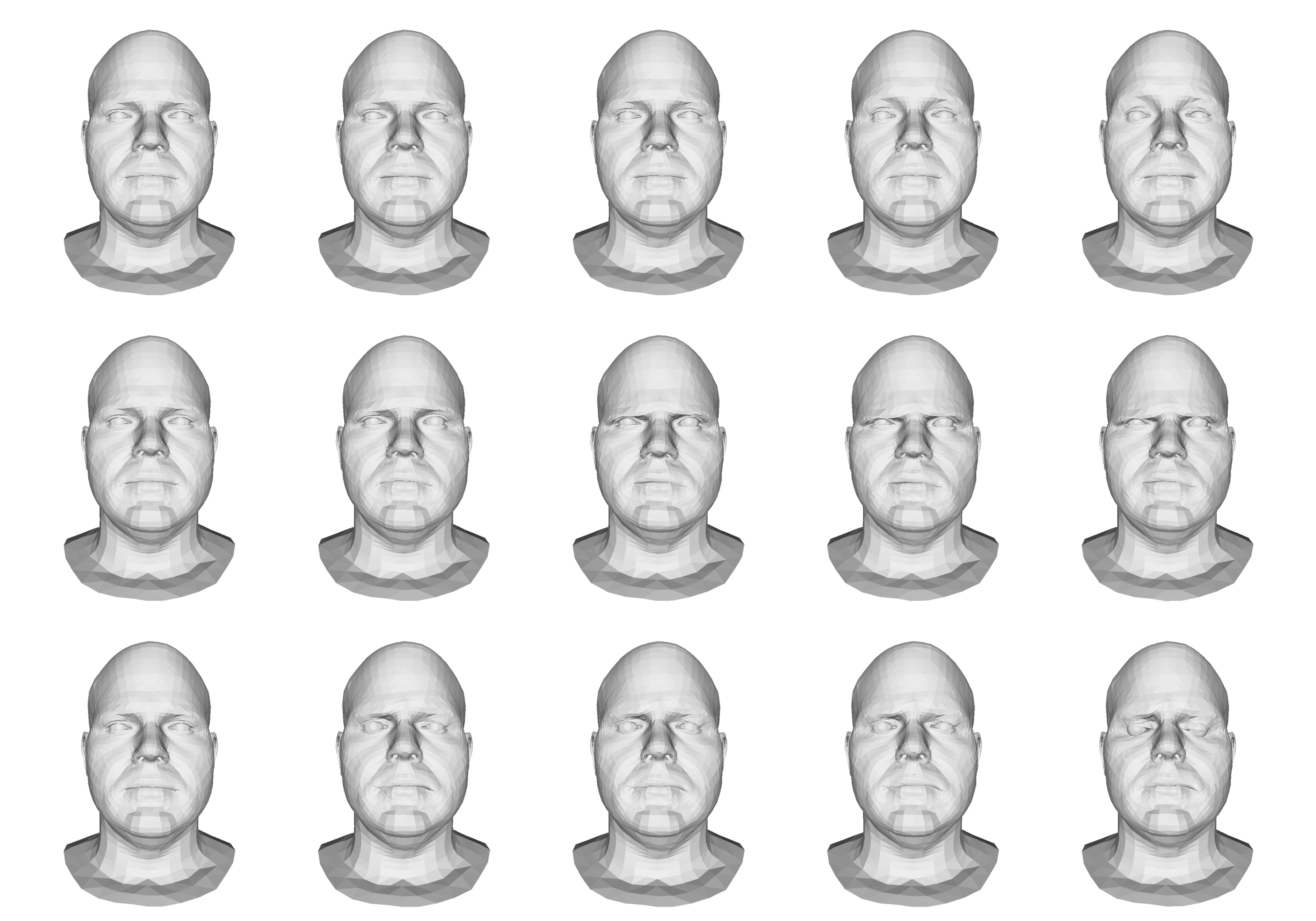}
   \end{minipage}
   \hfill\vline\hfill
   \begin{minipage}{.45\textwidth}
  \includegraphics[width=1\linewidth]{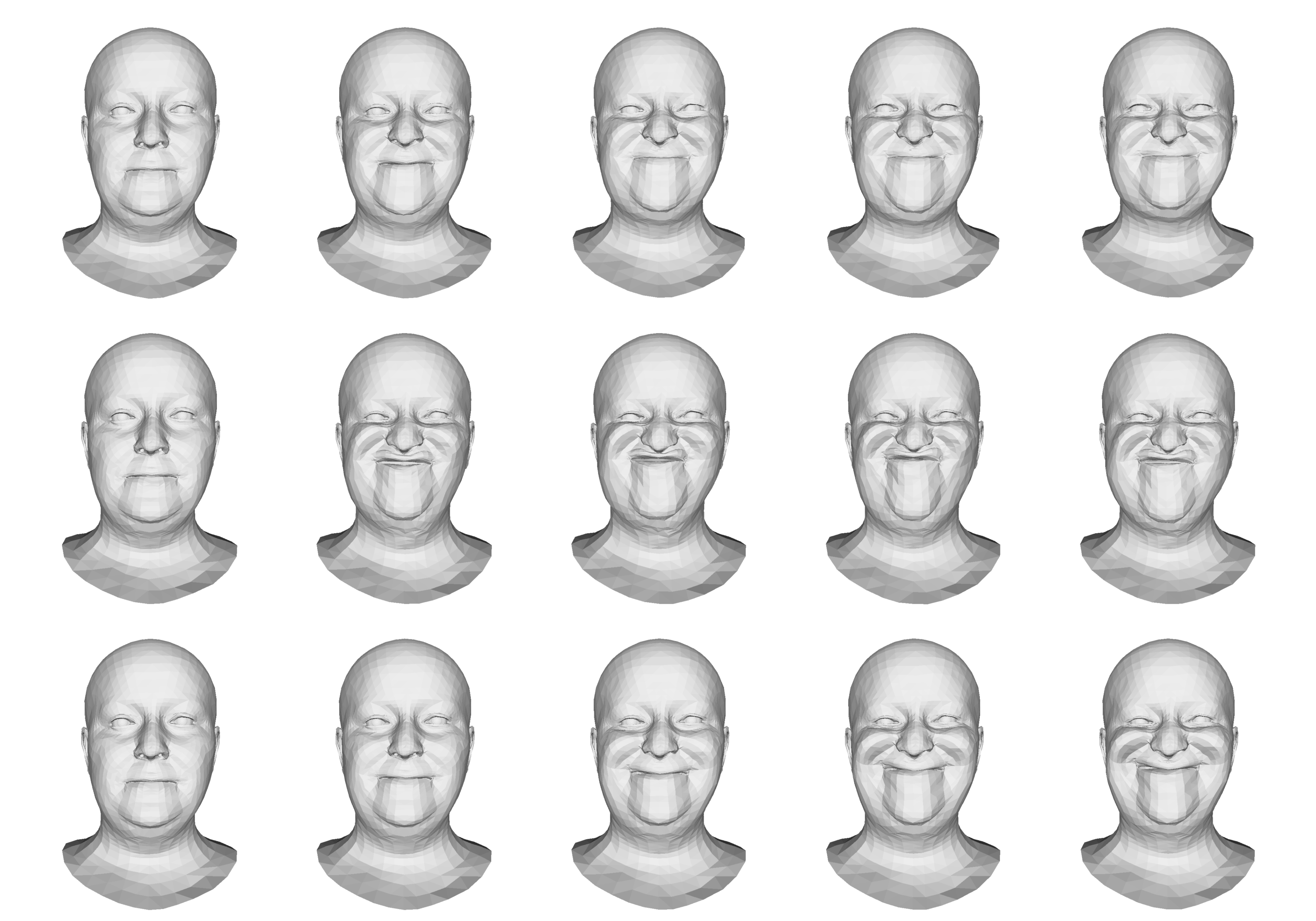}
  \end{minipage}
   \caption{
   Diversity of expressions generated with the label ``eyebrow''  (left), and ``high smile'' (right) in the geometry-adaptive generation task.
   All illustrated sequences are of type N2E.
   Note that eyebrows can be either lowered (the second and third rows) or raised (the first row). 
   Although the poses of maximal expression intensity look all similar in the three sequences of ``high smile'', their temporal properties are significantly different. %
   }
   \label{fig:exp_diversity2}
\end{figure}

Fig. \ref{fig:exp_diversity2} illustrates the variety of expressions we thus obtained
by using a same facial anatomy $L_M$ and a same label $y$ (either ``eyebrow'' or ``high smile''), which confirms that the proposed approach is able to generate expression sequences with sufficient level of diversity, even if a same facial anatomy is used for conditioning.

\section{Results related to landmark-guided mesh deformation}
To the best of our knowledge, only \cite{naima} and our work estimate $M_f$ from $M$ and $\Delta L_f$. Note that both approaches use spiral convolution. For the comparative experiments, we also adapt two autoencoders: CoMA \cite{coma}, which uses Chebyshev convolution and a mesh pooling, and the autoencoder proposed in \cite{casas2018learning} (the encoder and decoder consisting of three layers of linear, nonlinear, and linear activation units, respectively).
Both decoders, which originally take the latent representation of the input mesh as input, have been modified so as to consume the concatenation of the latent representation with $\Delta L_f$.%

We conducted two series of experiments: Either 3 expressions (expression split) or 3 subjects (identity split) have been excluded from the training set, and the performance of the model is evaluated on the excluded data. 
The mean per-vertex Euclidean error between the generated meshes and their ground truth has been measured to assess the performance.

Quantitative results are shown in Table \ref{tab:recon}. %
While the three methods based on spiral convolution %
generally yield effective results, our approach outperforms the others, thus confirming the advantage of the cross-attention layer, in particular. 

\begin{table}[htb!]

\caption{Per-vertex reconstruction error (mm).}
\centering

\begin{tabular}{lll}
\hline

Method & Expression split   & Identity split\\
\hline
Linear \cite{casas2018learning}& $0.67\pm0.76$    &  $0.73\pm 0.77$ \\
CoMA\cite{coma} &  $0.58\pm0.63$  &  $0.63\pm0.67$   \\
S2D \cite{naima} & $0.52\pm0.59$ & $0.55\pm0.62$   \\

Ours(w/o attention) & $0.54 \pm 0.59 $& 	$0.57\pm 0.64 $\\
Ours&  \textbf{$0.45 \pm 0.51$}&\textbf{$0.50\pm 0.58$ }\\

\noalign{\smallskip}

\hline
\end{tabular}
\label{tab:recon}
\end{table}

We propose to complement our quantitative analysis by a qualitative comparison of the different methods. 
As the "Expression split" and "Identity split" experiments yield very similar results, we focus solely on the "Identity split" experiment in the following.

\begin{figure}[h]
\begin{center}
   \includegraphics[width=0.6\linewidth]{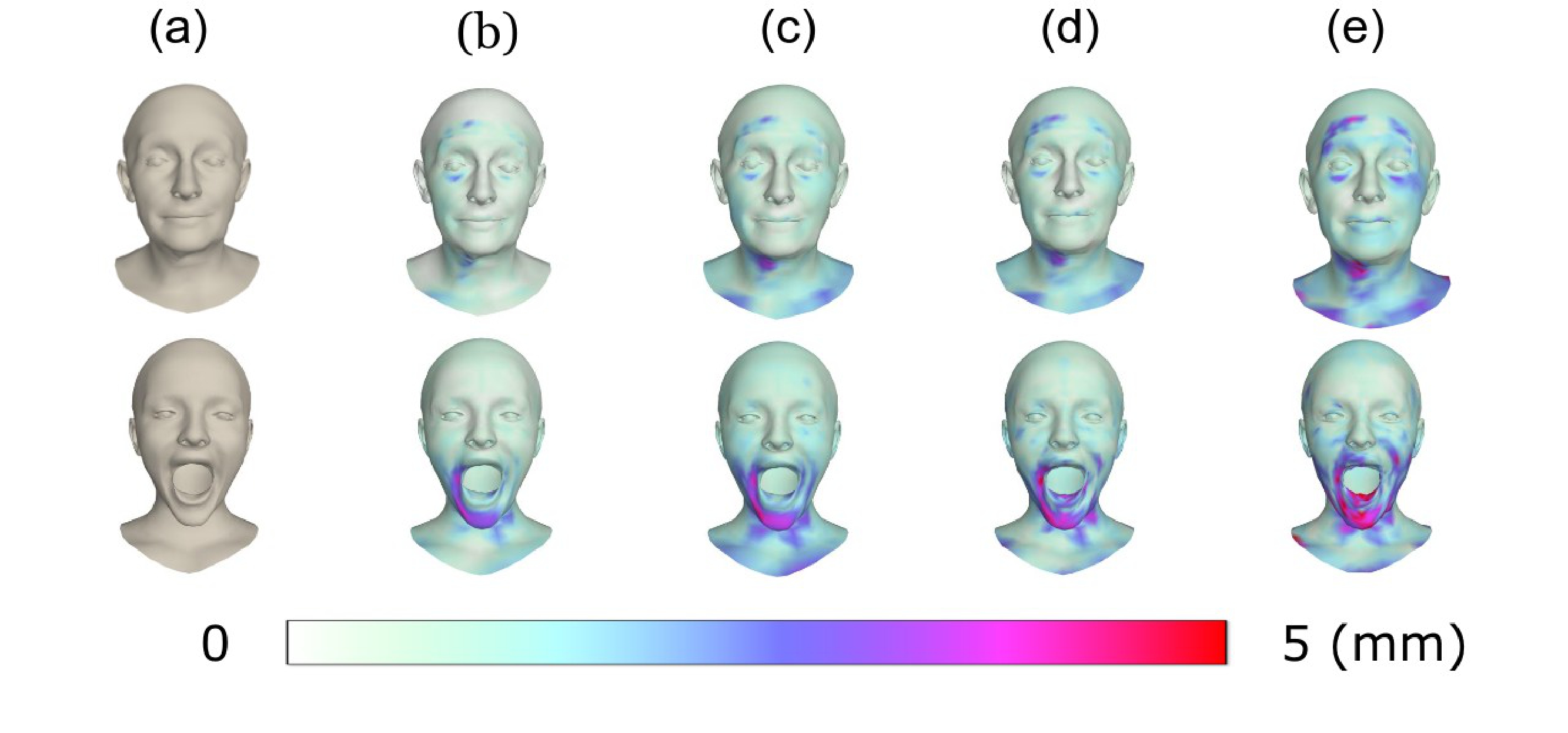}
\end{center}
   \caption{
   Qualitative comparison of our method (b) with S2D (c), CoMA (d), and Linear (e) in 
   the landmark-guided deformation of a given mesh.
   The ground truth meshes are given in the first column (a). The expression of the first row is close to the neutral face and that of the second row is taken from a sequence labeled as ``mouth extreme''.
   }
\label{fig:rec}
\setlength{\abovecaptionskip}{0.cm}
\end{figure}

Fig. \ref{fig:rec} depicts the ground truth mesh (a) 
as well as the meshes generated with several approaches (b-e). 
Each vertex of a generated mesh %
is assigned a color representing the Euclidean distance %
to its counterpart on the ground truth mesh. %
As expected, the errors appear mainly on the regions that have been 
deformed to attain the expression.
In Fig. \ref{fig:rec}, targeting an expression close to the neutral pose (first row) leads to tiny errors, whereas targeting an expression ``mouth extreme'' leads to errors that are mostly located near the mouth.
Our approach achieves the best performance in this qualitative error measure, confirming the quantitative results described above. %

Here we were able to deploy the reconstruction error as evaluation metric, since $\Delta L_f$ and $M$ pertain to the same individual. 
Besides, our encoder-decoder is also capable of retargeting landmark sequences to different facial meshes, a task previously shown by S2D \cite{naima}. As demonstrated in the videos on the project website, it produces results that are visually pleasing and qualitatively comparable to S2D. However, defining an evaluation metric for this task remains a challenging future endeavor.

\section{Conclusion}

We have presented a generator model to synthesize 3D dynamic facial expressions. The dynamics of facial expressions is first learned unconditionally, from which a series of downstream tasks 
are developed to synthesize an expression sequence conditioned on various condition signals. Also proposed is a robust face deformation scheme guided by the landmark set, which contributes to a higher reconstruction validity. Experimental results show that the proposed method can produce plausible face meshes of diverse types of expressions on different subjects. In addition, it outperforms SOTA models both qualitatively and quantitatively.
As has been demonstrated, our expression generation framework is versatile and can be used in many application scenarios including, but not limited to, label-guided generation, text-driven generation, geometry-adaptive generation, or expression filling.

\begin{acks}
This work was funded by the TOPACS ANR-19-CE45-0015 project, the binational project 
``Synthetic Data Generation and Sim-to-Real Adaptive Learning for Real-World Human Daily Activity Recognition of Human-Care Robots (21YS2900)" granted by ETRI, South Korea, and Human4D ANR-19-CE23-0020 project funded by the French National Research Agency (ANR).

\end{acks}

\bibliographystyle{ACM-Reference-Format}
\bibliography{main}

\appendix

%
%
\section{Training with and generation of sequences of arbitrary length}
\label{length}
Since our noise approximator is a bidirectional transformer, it can take sequences of arbitrary length as input \textemdash It can be trained using sequences of any length, and we can sample from the resulting model to obtain sequences of desired lengths (The length of $x_0$ will be that of $x_T$). 
In the same way, as a bidirectional transformer is used also to guide the reverse process, 
it can guide the reverse process with any length for $x_t$.
Consequently, tasks related to label control, text control, and geometry-adaptive generation can generate sequences of any desired length. %
Furthermore, the sequences that have to be filled with the expression filling task can be of any length.
For the sake of simplicity, 
we describe here only the label control task. 
The noise approximator and the classifier used for the guidance are either trained using sequences of a fixed length ($F=40$) or variable lengths ($F$ is uniformly distributed in the interval $[35,45]$). 

\begin{figure}[h]
  \centering
  \begin{minipage}{.49\textwidth}
  
  \includegraphics[width=0.8\linewidth]{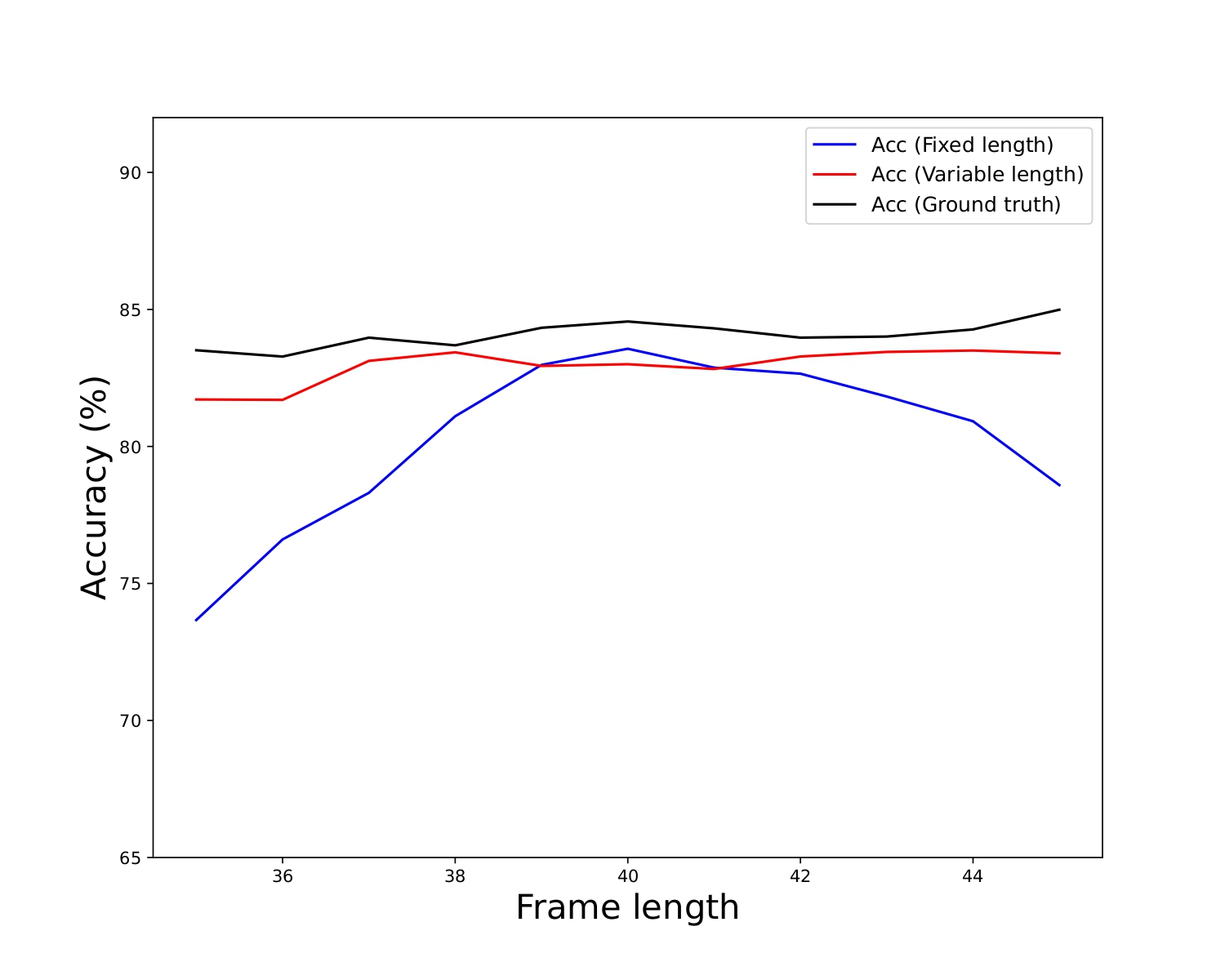}

  \end{minipage}
  \hfill\hfill
   \begin{minipage}{.49\textwidth}
  \includegraphics[width=0.8\linewidth]{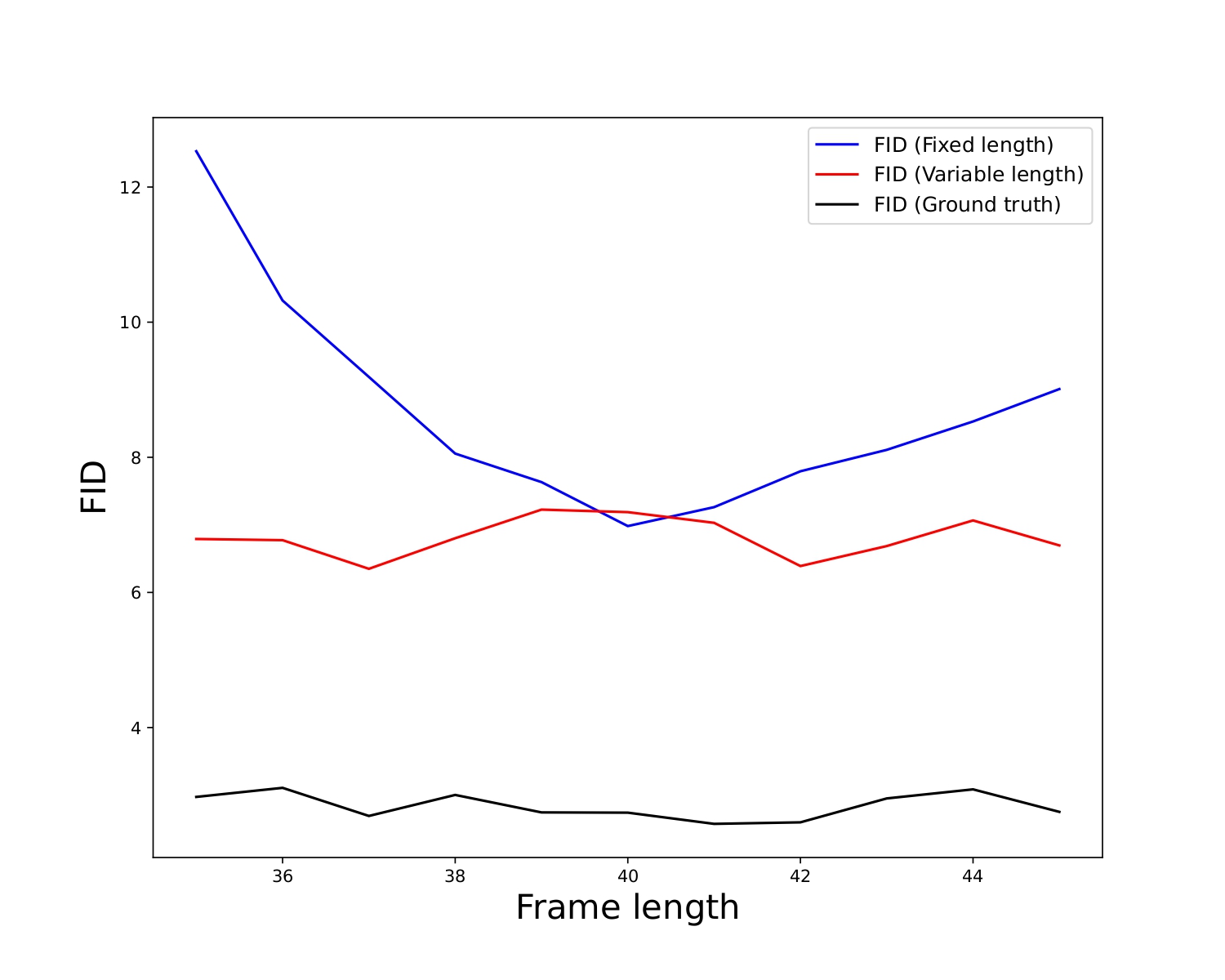}
  \end{minipage}
   \caption{Quantitative evaluation of the label control task for models trained with sequences of a fixed length ($F=40$) or  variable lengths. Performance is evaluated %
  on generated sequences of different lengths using, as in Sec. \ref{sub:sec:labelcontrol}, the classification accuracy (left) and the FID score (right).}
   \label{fig:length}
\end{figure}

The performance of both models is evaluated when outputting sequences of length in  $[35,45]$. 
The performance is evaluated as in Sec. 
\ref{sub:sec:labelcontrol}, except that the independent classifier is trained with sequences of variable length ($F$ is uniformly distributed in $[35,45]$).
Results are shown in Fig.\ref{fig:length}.

When generating sequences of different lengths is required, training with variable lengths helps the model to perform better. Moreover, the results obtained with the model trained with sequences of variable length are satisfactory: the achieved accuracy is similar to that of the ground truth. Moreover, the FID obtained for a length frame of 40 is similar to that calculated with the model dedicated to output sequences of length 40.

\section{Diversity of the generated sequences when conditioning on expression label}
\label{sec:diversity}

We study in this section the diversity of the generated sequences both in terms of facial anatomy ($L$) and in terms of expression ($\Delta L_f$) in the label control task. %
As a reminder, the 3D arrangement of a landmark frame $L_f$ can be regarded as the combination of the facial anatomy (at a neutral pose $L$) and the expression-driven shape change applied to it, i.e. $L_f=\Delta L_f + L$.

Since the proposed landmark-guided mesh deformation retargets the expression $\Delta L_f = L_f -L$ onto a new face anatomy given as a mesh $M$, it is used hereafter to illustrate the diversity of the generated expressions but it is not adapted to analyze the facial anatomy of the generated $L$. %
To show the diversity of facial anatomy generated by our model, we use the FLAME model\cite{flame} to compute the facial mesh from the landmark set of neutral pose\footnote{We can note that the meshes generated from FLAME lack certain details of the facial geometry, resulting in dull, lifeless
shapes. 
Furthermore, FLAME takes about $470$s to fit one sequence, while the proposed landmark-guided mesh deformation needs only about $1.30$s.}.

\begin{figure}[h]
  \centering

  \includegraphics[width=0.3\linewidth]{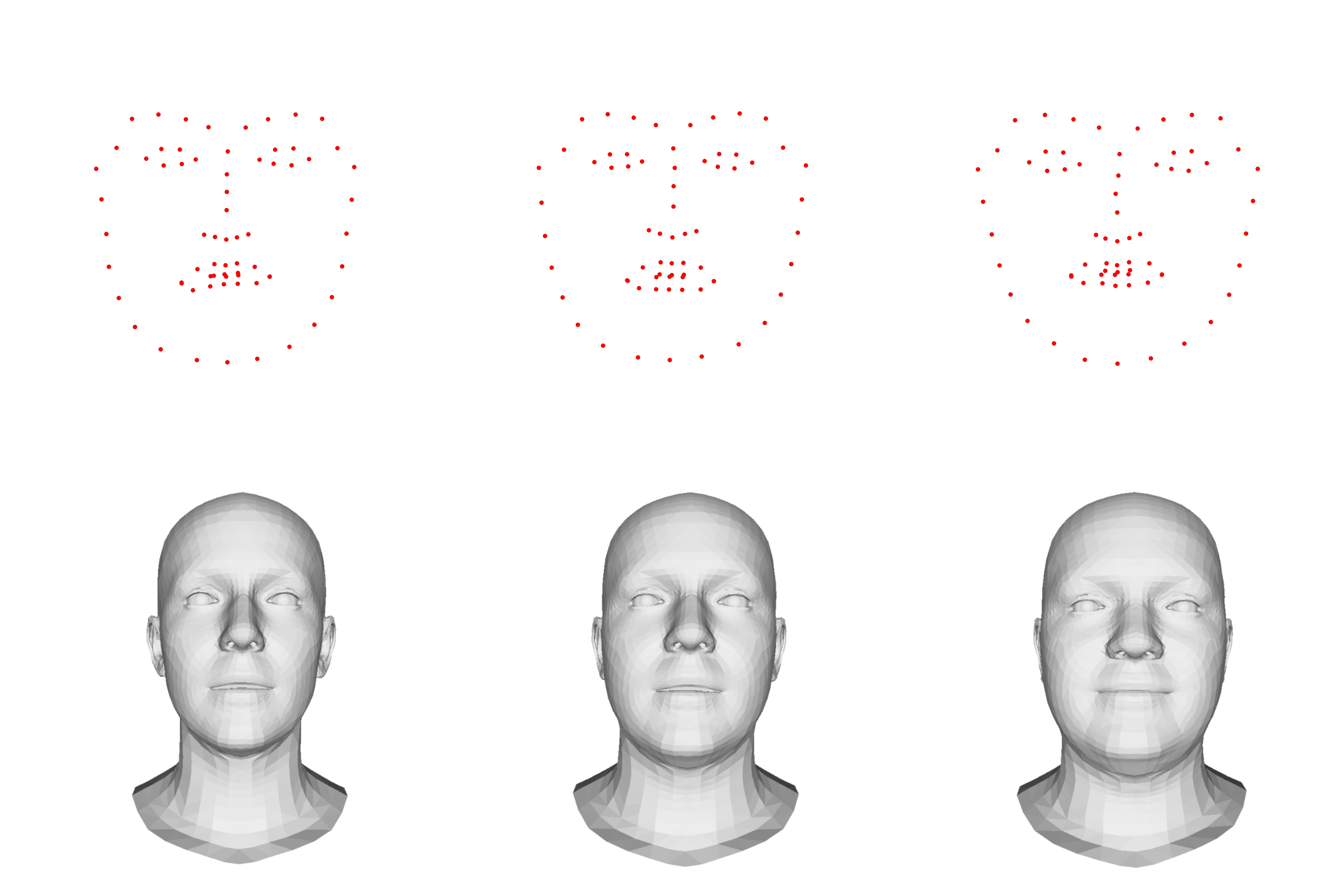}
   \caption{
   Diversity of facial anatomy in the generated expressions. We use FLAME model to compute facial meshes from the landmark sets, for the visualization purpose.
   }
   \label{fig:shape_diversity}
\end{figure}

\begin{figure}[h]
  \centering
  \begin{minipage}{.45\textwidth}

  \includegraphics[width=1\linewidth]
  {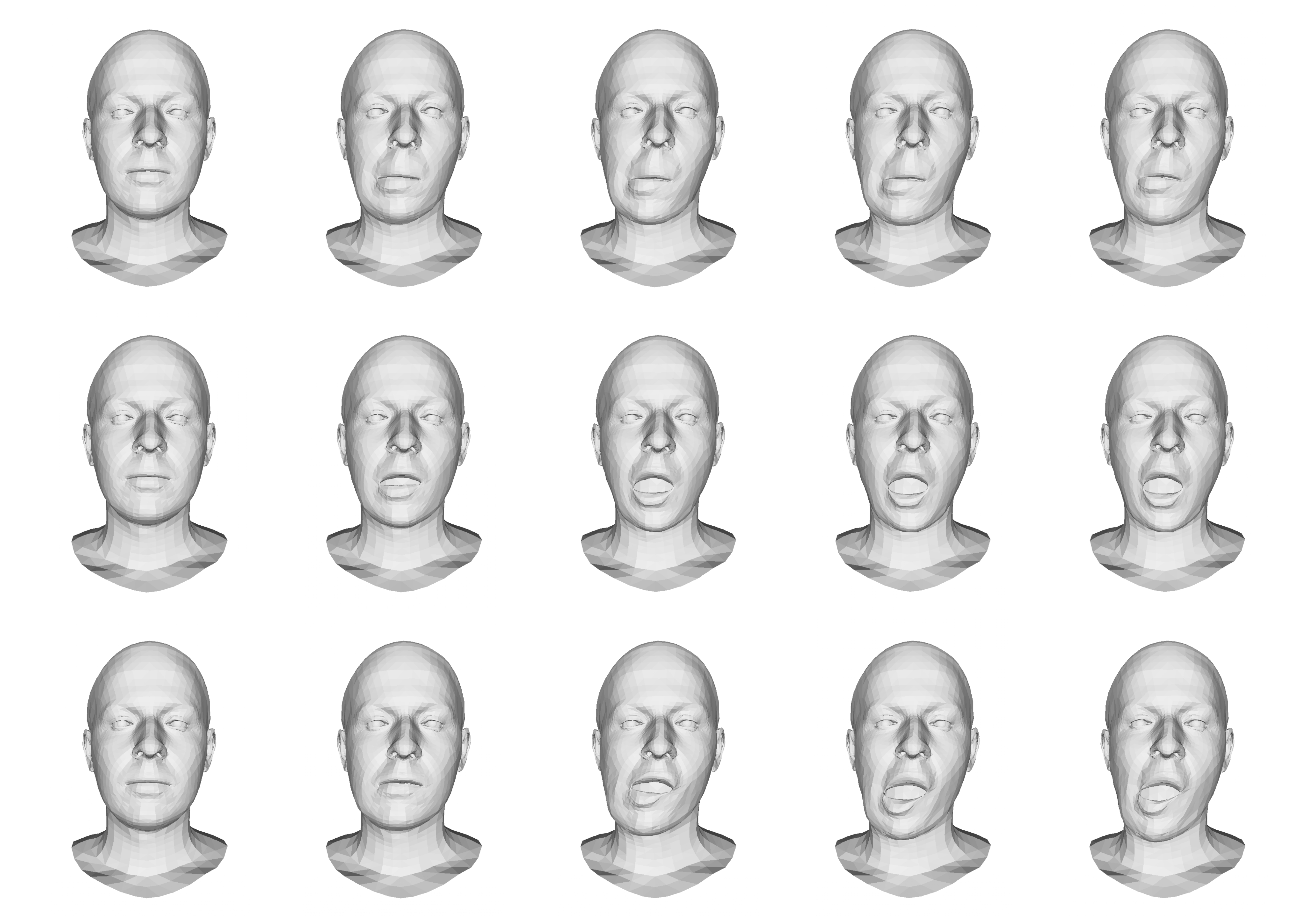}

  \end{minipage}
  \hfill\vline\hfill
   \begin{minipage}{.45\textwidth}
  \includegraphics[width=1\linewidth]{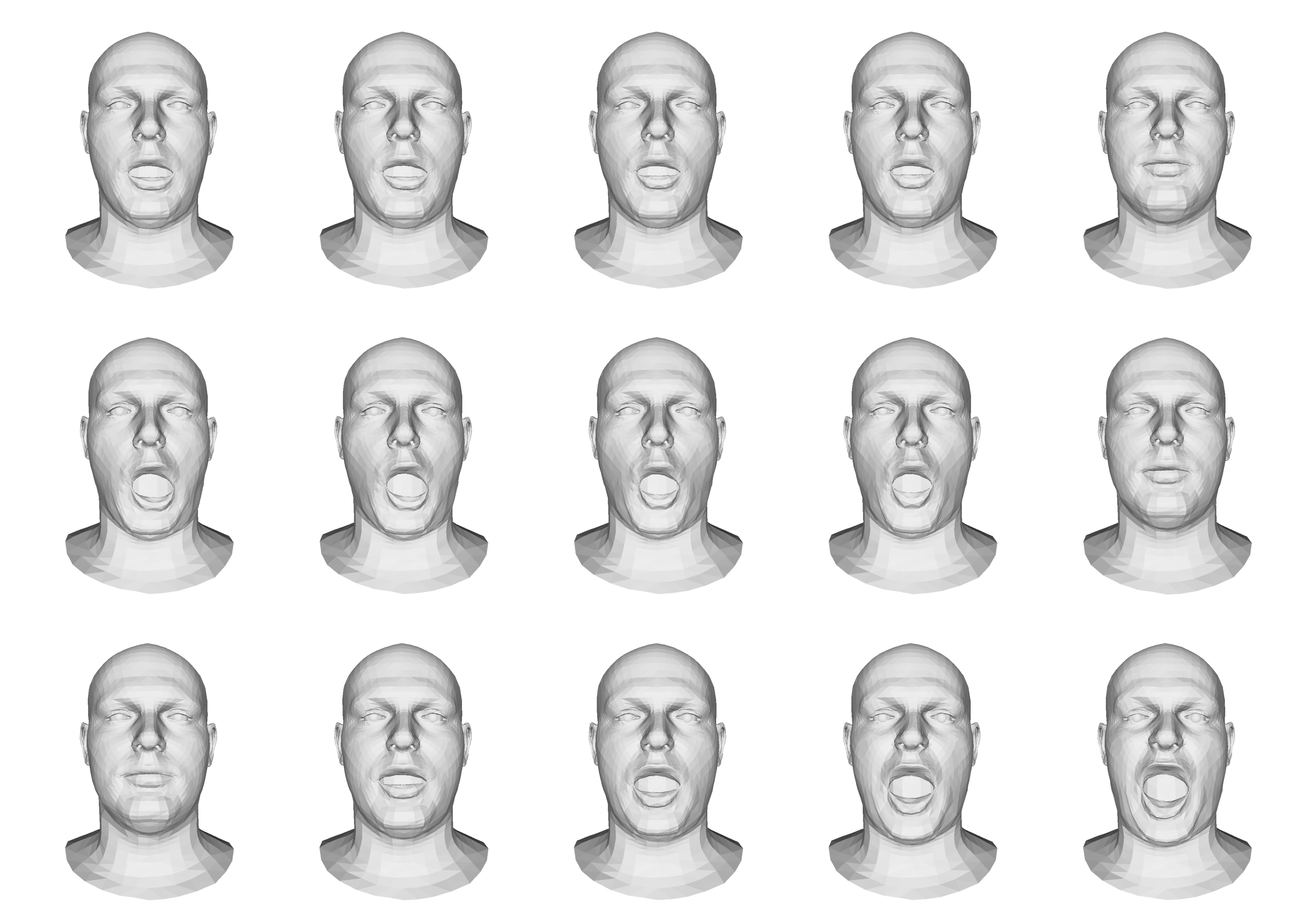}
  \end{minipage}
   \caption{
   Diversity of expressions generated with the label ``mouth side'' (left), and ``mouth open'' (right)  in the label control task. Note that generated sequences can be either of type E2N or N2E.}
   \label{fig:exp_diversity1}
\end{figure}

Fig. \ref{fig:shape_diversity} presents %
three illustrative neutral faces $L$ that we generated by conditioning the reverse process on the same expression label ``mouth open''. Both landmark set and the FLAME-fitted mesh are shown,                                        for each face.  
(The neutral face $L$ associated with a generated sequence $x_0$ is set to either $L_1$ or $L_F$, depending on the sequence type.) 
Additionally, the diversity in the generated expression is illustrated in Fig. \ref{fig:exp_diversity1}. The apparent distinction among these results demonstrate that the proposed approach is able to generate sequences of rich diversity, both in terms of facial anatomy and expression (This is due to the input noise $x_{T}$ that is sampled from $\mathcal{N}(0,I)$).

\section{Pseudo code for each downstream task}
\label{Pseudocode}

\begin{algorithm}[H]
\tiny
\caption{Label control}\label{algo:label}
\begin{flushleft} 
\hspace*{\algorithmicindent} \textbf{Input}: Label $y$. \\
\hspace*{\algorithmicindent} \textbf{Output}: Sequence $x_0$ (corresponding to label $y$). 
\end{flushleft} 
\begin{algorithmic}[1]

\State $x_T \sim N(0, I) $ 

\For {\texttt{$t=T,...,1$ }} 

\State \algorithmiccomment{Estimation of $p_{\theta}(.|x_t)$}

\State Compute ${\epsilon}_\theta\left(x_t, t\right)$

\State Compute $\mu_\theta(x_t, t)$: $\mu_\theta(x_t, t)=\frac{1}{\sqrt{\alpha_t}}\left(x_t-\frac{\beta_t}{\sqrt{1-\bar{\alpha}_t}} {\epsilon}_\theta\left(x_t, t\right)\right)$ 

\State \algorithmiccomment{Sampling from $p_{\theta}(.|x_t)$}

\State $z \sim N(0, I)$ if $t>1$, $0$ otherwise 

\State Set $\hat{x}_{t-1}$ to $\mu_\theta(x_t, t)+\sigma_t z$

\State \algorithmiccomment{Optimization: optimization procedure is initialized with $\hat{x}_{t-1}$}

\State $x_{t-1} = \argmax \limits_{x}  \left[\lambda log(p_\theta(x|x_t))+log(p_\phi(y|x,t-1)) \right]$

\EndFor
\Return $x_0$
\end{algorithmic}
\end{algorithm}

\begin{algorithm}[H]
\tiny
\caption{Text control}\label{algo:text}
\begin{flushleft} 
\hspace*{\algorithmicindent} \textbf{Input}: Text $c$. \\
\hspace*{\algorithmicindent} \textbf{Output}: Sequence $x_0$ (corresponding to text $c$). 
\end{flushleft} 
\begin{algorithmic}[1]

\State $x_T \sim N(0, I) $ 

\For {\texttt{$t=T,...,1$ }} 

\State \algorithmiccomment{Estimation of $p_{\theta}(.|x_t)$}

\State Compute ${\epsilon}_\theta\left(x_t, t\right)$

\State Compute $\mu_\theta(x_t, t)$: $\mu_\theta(x_t, t)=\frac{1}{\sqrt{\alpha_t}}\left(x_t-\frac{\beta_t}{\sqrt{1-\bar{\alpha}_t}} {\epsilon}_\theta\left(x_t, t\right)\right)$ 

\State \algorithmiccomment{Sampling from $p_{\theta}(.|x_t)$}

\State $z \sim N(0, I)$ if $t>1$, $0$ otherwise 

\State Set $\hat{x}_{t-1}$ to $\mu_\theta(x_t, t)+\sigma_t z$

\State \algorithmiccomment{Optimization: optimization procedure is initialized with $\hat{x}_{t-1}$}

\State  $x_{t-1} = \argmax \limits_{x} \left[\lambda log(p_\theta(x|x_t))+ cos(BiT(x, t-1), CLIP(c))\right]$

\EndFor
\Return $x_0$
\end{algorithmic}
\end{algorithm}

\begin{algorithm}[H]
\tiny
\caption{Sequence filling}\label{algo:filling}
\begin{flushleft} 
\hspace*{\algorithmicindent} \textbf{Input}: Partial sequence $x_0|_{\mathcal{S}_K}$ \\
\hspace*{\algorithmicindent} \textbf{Output}: Completed sequence $x_0$ 
\end{flushleft} 
\begin{algorithmic}[1]

\State $x_T|_{\mathcal{S}_U} \sim N(0, I)$
\State $x_T|_{\mathcal{S}_K} = \sqrt{\bar{\alpha}_{T}} x_0|_{S_k} +\sqrt{1-\bar{\alpha}_
{T}} \epsilon, \epsilon \sim N(0, I)$

\For {\texttt{$t=T,...,1$ }} 

\State \algorithmiccomment{Estimation of $p_{\theta}(.|x_t)$}

\State Compute ${\epsilon}_\theta\left(x_t, t\right)$

\State Compute $\mu_\theta(x_t, t)$: $\mu_\theta(x_t, t)=\frac{1}{\sqrt{\alpha_t}}\left(x_t-\frac{\beta_t}{\sqrt{1-\bar{\alpha}_t}} {\epsilon}_\theta\left(x_t, t\right)\right)$ 

\State \algorithmiccomment{Sampling from $p_{\theta}(.|x_t)$}

\State $z \sim N(0, I)$ if $t>1$, $0$ otherwise 

\State Set $\hat{x}_{t-1}$ to $\mu_\theta(x_t, t)+\sigma_t z$

\State ${x_{t-1}}|_{\mathcal{S}_U}$ = ${\hat{x}_{t-1}}|_{\mathcal{S}_U}$

\If {$t>1$} \algorithmiccomment{if $t=1$, ${x_{0}}|_{\mathcal{S}_K}$ is already properly set}.
    \State $x_{t-1}|_{S_K} = \sqrt{\bar{\alpha}_{t-1}} x_0|_{S_k} +\sqrt{1-\bar{\alpha}_{t-1}} \epsilon, \epsilon \sim N(0, I)$
    \EndIf

\EndFor
\Return $x_0$
\end{algorithmic}
\end{algorithm}

\begin{algorithm}[H]
\tiny
\caption{Geometry-adaptive generation with label control}\label{algo:geo}
\begin{flushleft} 
\hspace*{\algorithmicindent} \textbf{Input}: Label $y$ and partial sequence $x_0|_{\mathcal{S}_K}$.  $\mathcal{S}_K$ is either $\{1\}$ or $\{F\}$ and the unique frame associated with  $x_0|_{\mathcal{S}_K}$ is a neutral one.\\
\hspace*{\algorithmicindent} \textbf{Output}: Completed sequence $x_0$ (corresponding to label $y$)
\end{flushleft} 
\begin{algorithmic}[1]

\For {\texttt{$i = 1$ to $5$ }}

\If {i == 1}
\State $x_T|_{\mathcal{S}_U} \sim N(0, I)$
\State $x_T|_{\mathcal{S}_K} = \sqrt{\bar{\alpha}_{T}} x_0|_{S_k} +\sqrt{1-\bar{\alpha}_
{T}} \epsilon, \epsilon \sim N(0, I)$
\Else
\State $x_T = \sqrt{\bar{\alpha}_{T}} x_0 +\sqrt{1-\bar{\alpha}_{T}} \epsilon, \epsilon \sim N(0, I)$
\EndIf

\For {\texttt{$t=T,...,1$ }} 

\State \algorithmiccomment{Estimation of $p_{\theta}(.|x_t)$}

\State Compute ${\epsilon}_\theta\left(x_t, t\right)$

\State Compute $\mu_\theta(x_t, t)$: $\mu_\theta(x_t, t)=\frac{1}{\sqrt{\alpha_t}}\left(x_t-\frac{\beta_t}{\sqrt{1-\bar{\alpha}_t}} {\epsilon}_\theta\left(x_t, t\right)\right)$ 

\State \algorithmiccomment{Sampling from $p_{\theta}(.|x_t)$}

\State $z \sim N(0, I)$ if $t>1$, $0$ otherwise 

\State Set $\hat{x}_{t-1}$ to $\mu_\theta(x_t, t)+\sigma_t z$

\State \algorithmiccomment{Optimization: optimization procedure is initialized with $\hat{x}_{t-1}$}

\State $\hat{x}_{t-1} = \argmax \limits_{x}  \left[\lambda log(p_\theta(x|x_t))+log(p_\phi(y|x,t-1)) \right]$

\State ${x_{t-1}}|_{\mathcal{S}_U}$ = ${\hat{x}_{t-1}}|_{\mathcal{S}_U}$
\If {$t>1$} \algorithmiccomment{if $t=1$, ${x_{0}}|_{\mathcal{S}_K}$ is already properly set}.
    \State $x_{t-1}|_{S_K} = \sqrt{\bar{\alpha}_{t-1}} x_0|_{S_k} +\sqrt{1-\bar{\alpha}_{t-1}} \epsilon, \epsilon \sim N(0, I)$
    \EndIf

\EndFor
\EndFor
\Return $x_0$
\end{algorithmic}
\end{algorithm}

\end{document}